\documentclass[acmtog]{acmart}
\usepackage{booktabs}
\usepackage{graphicx}
\usepackage{pifont}
\usepackage{subcaption}
\usepackage{longtable}
\usepackage{tabularx}
\usepackage{array}
\usepackage{makecell}
\usepackage{caption}
\usepackage[export]{adjustbox}
\usepackage{ltablex}
\keepXColumns
\usepackage{xltabular}
\usepackage{placeins}
\usepackage{pdfpages}

\usepackage{kotex}

\usepackage[ruled]{algorithm2e}

\SetAlFnt{\small}
\SetAlCapFnt{\small}
\SetAlCapNameFnt{\small}
\SetAlCapHSkip{0pt}

\newcommand{\cmark}{\ding{51}}
\newcommand{\xmark}{\ding{55}}

\acmJournal{TOG}

\setcopyright{rightsretained}   % 저자가 권리 보유 (프리프린트에 적합)
\renewcommand\footnotetextcopyrightpermission[1]{}

\begin{document}
% \pagestyle{plain}

% ---------------------------------------------------------------
% Title

\title{Look Before You Edit: Attention-Guided Camera Placement and Multi-View Alignment for 3D Gaussian Splatting Editing}

% ---------------------------------------------------------------
% Author information
% For anonymous SIGGRAPH technical paper submissions, do not enter author information.

\author{Jaeyeon Park}
\orcid{0009-0002-0952-7573}
% \authornote{Corresponding author.}
\affiliation{%
  \institution{Seoul National University}
  \city{Seoul}
  \country{South Korea}
}
\email{jypark@snu.ac.kr}

\author{Taeho Kang}
\orcid{0000-0002-4556-5588}
\affiliation{%
  \institution{Seoul National University}
  \city{Seoul}
  \country{South Korea}
}
\email{taeho.kang@hcs.snu.ac.kr}

\author{Youngki Lee}
\orcid{0000-0002-1319-7071}
\affiliation{%
  \institution{Seoul National University}
  \city{Seoul}
  \country{South Korea}
}
\email{youngki.lee@gmail.com}

\renewcommand\shortauthors{Park et al.}

% ---------------------------------------------------------------
% Abstract
% IMPORTANT:
% In acmart/SIGGRAPH, abstract must appear BEFORE \maketitle.
% If sections/00_abstract.tex already contains:
%   \begin{abstract}
%   ...
%   \end{abstract}
% then the line below is okay.
%
% If sections/00_abstract.tex contains only abstract text without the environment,
% replace this with:
%
% \begin{abstract}
% \input{sections/00_abstract}
% \end{abstract}

\begin{abstract}
Text-driven 3D scene editing with 3D Gaussian Splatting (3DGS) typically applies a 2D diffusion editor to views rendered from fixed training cameras, limiting both the spatial coverage of edits and the user's freedom to target specific objects in complex scenes.
We present \textbf{LB-Edit}, a framework that addresses two coupled problems: \emph{where to place editing cameras} for localized edits, and \emph{how to make per-view edits agree with one another} so that the 3D scene remains consistent after fine-tuning.
First, Attention-Guided Editing Camera Placement (\textbf{ACP}) probes the diffusion model's self- and cross-attention at multiple candidate camera distances to find where attention is well-contained in the region of interest, then places a compact, geometrically diverse editing camera set at that attention-optimal distance.
Second, Multi-view Attention Alignment (\textbf{MAA}) steers the editor toward the same edit across views along two axes: it aligns \emph{appearance} by sharing self-attention features via token-level correspondence, and aligns \emph{spatial location} by lifting cross-attention maps onto the 3D Gaussians as a shared 3D attention field, suppressing both appearance and spatial drift.
Experiments on multi-object and single-object scenes show that our method achieves the highest user preference in instruction fidelity, multi-view consistency, and editing locality, using as few as 5 editing views and reducing latency by up to $7\times$ over existing methods.
\end{abstract}

% ---------------------------------------------------------------
% CCS Concepts
% TODO: Replace these with proper CCS concepts from https://dl.acm.org/ccs



% ---------------------------------------------------------------
% Keywords


% ---------------------------------------------------------------
% Make title

% ---------------------------------------------------------------
% Main content

\begin{teaserfigure}
    \centering
    \includegraphics[width=\textwidth]{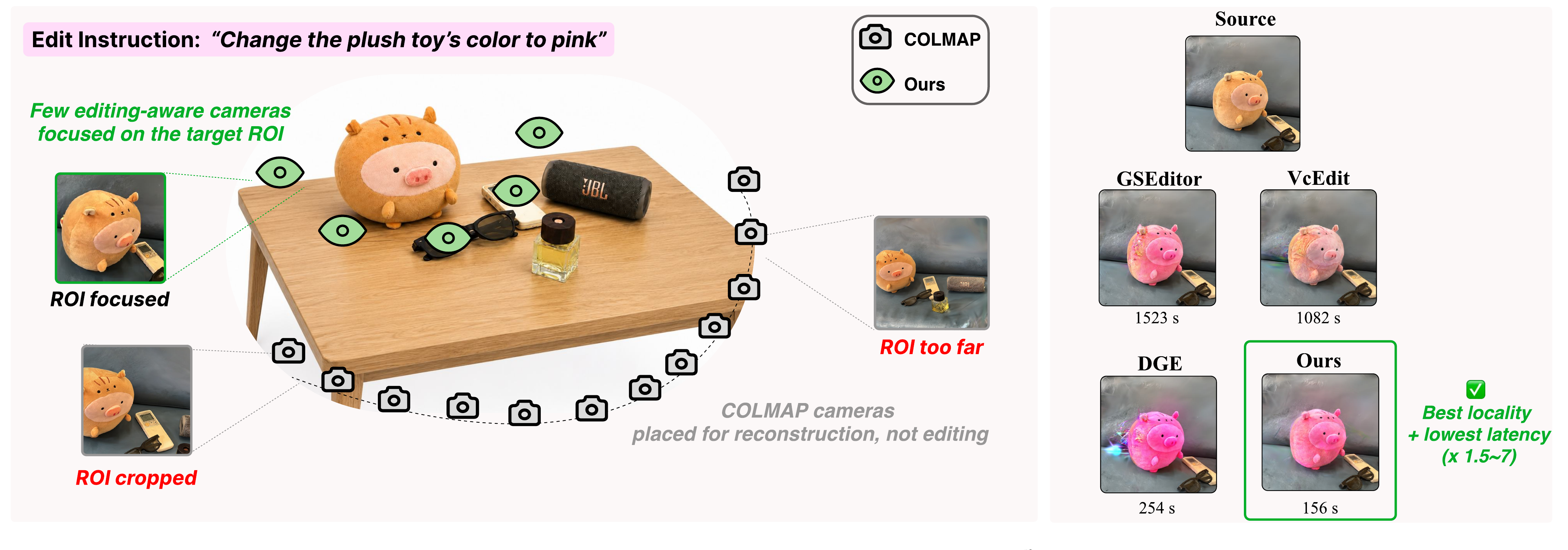}
    \caption{
    \textbf{LB-Edit.}
    COLMAP cameras are optimized for reconstruction, not editing: target objects may
    be cropped or observed from views that are too distant for reliable attention
    localization. Given a text-specified ROI, our method selects a compact set of
    editing-aware views at attention-effective distances, enabling more localized
    3DGS edits with substantially lower latency than prior methods.
    }
    \Description{
    A teaser figure comparing COLMAP reconstruction cameras with editing-aware cameras selected by the proposed method.
    }
    \label{fig:teaser}
\end{teaserfigure}
 % teaser figure

\maketitle
\thispagestyle{plain}   % 첫 페이지 footer 제거
\pagestyle{plain}       % 이후 페이지 footer 제거

\section{Introduction}
\label{sec:intro}
3D Gaussian Splatting (3DGS)~\cite{kerbl2023_3dgs} has emerged as a powerful alternative to Neural Radiance Fields (NeRF)~\cite{mildenhall2021nerf}, offering real-time rendering through an explicit representation of discrete Gaussian primitives.
Leveraging 2D diffusion priors~\cite{rombach2022high, brooks2023instructpix2pix}, recent methods have demonstrated promising text-driven 3DGS editing: GaussianEditor~\cite{chen2024gaussianeditor, wang2024gaussianeditor} introduced language-guided ROI localization, GaussCtrl~\cite{wu2024gaussctrl} and VcEdit~\cite{wang2024vcedit} improved multi-view consistency via attention sharing, and DGE~\cite{chen2024dge} achieved direct editing through extended self-attention inspired by TokenFlow~\cite{geyer2024tokenflow}.
A common thread across these methods is that they apply the 2D editor to views rendered from the \emph{fixed COLMAP training cameras}~\cite{schoenberger2016sfm} used during reconstruction.
While effective when the edit target is a single dominant object that occupies a large fraction of every training view, this design tightly couples editing supervision to a camera distribution that was optimized for reconstruction quality, not for editing a specific region.

In practical AR/VR settings, a user wants to freely select a specific object in a complex scene using natural language~\cite{liu2024groundingdino, kirillov2023segment} and edit only that region.
Once the target is no longer the dominant content of the scene, reusing the training cameras becomes problematic: the ROI may be cropped at the image border, partially occluded, or observed from distances at which the 2D editor cannot reliably localize the edit, regardless of how consistency is enforced downstream.
Placing new editing cameras near the ROI in free 3D space is a natural remedy, but introduces two coupled challenges that prior work does not address.
First, the diffusion model's editing behavior is highly sensitive to camera-to-object distance: too close and attention leaks across the entire image, too far and the edit fails to localize.
Second, independently editing each freshly placed view produces \emph{appearance drift} (inconsistent style/shape) and \emph{spatial drift} (misaligned edit locations) across views, and these per-view inconsistencies fuse into artifacts when the edited images are used to fine-tune the 3D scene.
Existing approaches address multi-view consistency via self-attention~\cite{chen2024dge} or cross-attention~\cite{wang2024vcedit} alignment in isolation, and none consider how the choice of editing cameras itself shapes the attention behavior that these alignment mechanisms operate on.

We propose \textbf{LB-Edit} (short for \emph{Look Before You Edit}), a two-stage framework that treats camera placement and multi-view consistency as a single, attention-aware problem.
\textbf{Attention-Guided Editing Camera Placement (ACP)} probes the diffusion model's self- and cross-attention at a few candidate camera distances to find where attention is well-contained in the ROI, then constructs a compact, geometrically diverse editing camera set at that attention-optimal distance.
\textbf{Multi-View Attention Alignment (MAA)} then steers the editor toward the same edit across the resulting views along two axes---appearance (self-attention) and spatial location (cross-attention)---by sharing self-attention features via token-level correspondence and lifting cross-attention maps onto the 3D Gaussians as a shared 3D attention field, suppressing both appearance and spatial drift.
Experiments on multi-object and single-object scenes show that our method achieves the highest user preference across instruction fidelity, multi-view consistency, and editing locality, while using as few as 5 editing cameras---reducing latency by up to $7\times$ compared to baselines that require 20--60 editing views.

Our contributions are:
\begin{itemize}
    \item \textbf{Attention-Guided Editing Camera Placement (ACP)}, a novel camera placement strategy for diffusion-based 3D editing that leverages the editor's own attention statistics to determine the attention-optimal editing distance and construct a diverse, ROI-tailored editing camera set.
    \item \textbf{Multi-View Attention Alignment (MAA)} that aligns the editor's behavior across views along two axes---appearance (self-attention) and spatial location (cross-attention)---via token-level feature matching and a shared 3D attention field.
    \item State-of-the-art editing quality and efficiency on complex multi-object scenes, with the highest user preference and competitive CLIP scores at significantly reduced computational cost.
\end{itemize}

\section{Related Work}
\label{sec:related}

\subsection{3D Editing with Text Prompts}
Text-driven 3D scene editing is typically built on top of powerful 2D diffusion editors~\cite{rombach2022high}.
InstructPix2Pix~\cite{brooks2023instructpix2pix} learns instruction-guided edits from paired data, while Prompt-to-Prompt~\cite{hertz2022prompt} and ControlNet~\cite{zhang2023controlnet} expose cross-attention and spatial conditioning as controllable primitives.
3D methods lift these capabilities to neural radiance fields or 3D Gaussian splats either by optimizing the 3D representation with score-distillation-sampling (SDS) losses~\cite{poole2023dreamfusion,wang2024gaussianeditor} or by repeatedly applying a 2D editor to rendered views and re-fitting the 3D model~\cite{haque2023instructnerf2nerf}.

\subsection{Camera Placement for 3D Editing}
The placement of editing cameras affects both quality and cost, yet most diffusion-based 3D editors simply reuse the COLMAP training cameras from reconstruction~\cite{haque2023instructnerf2nerf,wang2024gaussianeditor,chen2024dge,wu2024gaussctrl,wang2024vcedit}, which are placed for reconstruction quality rather than for editing a specific region.
A few methods select a smaller set of key views via hand-crafted heuristics: ViCA-NeRF~\cite{dong2023vicanerf} uses a modified-pixel ratio and propagates edits via depth-based warping, while trajectory- or frame-based variants~\cite{luo2024trame,khalid2024_3dego} order keyframes along a camera path.
More recently, user-interactive methods~\cite{wen2025intergsedit,huang2025edit360} let users designate a key or anchor view, and post-hoc approaches~\cite{qu2025editcast3d} filter edited frames by reconstruction quality.
None of these strategies examine how the diffusion editor itself behaves at a given camera placement.
Our Attention-Guided Editing Camera Placement (Sec.~\ref{sec:acp}) is, to our knowledge, the first to make editing-camera placement an attention-driven decision: we probe candidate camera distances using diffusion self- and cross-attention statistics and place a compact camera set at the distance where edits are best contained in the ROI (see the CP column in Table~\ref{tab:related_work_summary}).

\subsection{Consistency in Diffusion-based 3D Editing}

Applying 2D diffusion editors to 3D scenes often produces inconsistent supervision across views, which leads to artifacts when edited images are fused into a 3D representation.
Early methods address this with iterative optimization.
Instruct-NeRF2NeRF~\cite{haque2023instructnerf2nerf} alternates between per-view InstructPix2Pix edits and NeRF re-optimization, while GaussianEditor~\cite{wang2024gaussianeditor} adapts the idea to 3D Gaussian splatting with localized SDS-based editing.
However, these methods treat the 2D editor as a black box and do not regulate internal attention.

More recent work attempts to enforce consistency through attention alignment.
DGE~\cite{chen2024dge} and GaussCtrl~\cite{wu2024gaussctrl} align self-attention across views using spatio-temporal attention or cross-view latent attention, improving appearance consistency.
Conversely, VcEdit~\cite{wang2024vcedit} and InterGSEdit~\cite{wen2025intergsedit} enforce spatial consistency by lifting cross-attention maps into 3D and re-rendering them for each view.
As summarized in Table~\ref{tab:related_work_summary}, prior work aligns either self-attention or cross-attention but not both.
Our method jointly synchronizes both mechanisms to suppress appearance and spatial drift while additionally introducing attention-guided editing-camera placement.

% \begin{table}[t]
% \centering
% \small
% \setlength{\tabcolsep}{4pt}

% \caption{Comparison of camera placement and attention alignment strategies in diffusion-based 3D editing. CP = camera-placement strategy beyond fixed COLMAP cameras; SA = self-attention alignment across views; CA = cross-attention alignment across views.}
% \label{tab:sa_ca_comparison}

% \resizebox{\linewidth}{!}{
% \begin{tabular}{lcccl}
% \toprule
% Method & CP & SA & CA & Multi-View Consistency Strategy \\
% \midrule
% IN2N~\cite{haque2023instructnerf2nerf} & \xmark & \xmark & \xmark & Iterative dataset update \\
% GaussianEditor~\cite{wang2024gaussianeditor} & \xmark & \xmark & \xmark & Gaussian Semantic Tracing + HGS \\
% ViCA-NeRF~\cite{dong2023vicanerf} & Pixel-ratio key views & \xmark & \xmark & Depth-warped edit propagation \\
% DGE~\cite{chen2024dge} & \xmark & \cmark & \xmark & Extended ST self-attn + feature injection \\
% GaussCtrl~\cite{wu2024gaussctrl} & \xmark & \cmark & \xmark & Cross-view latent attn in attn1 \\
% VcEdit~\cite{wang2024vcedit} & \xmark & \xmark & \cmark & 3D inverse splatting of CA maps \\
% EditSplat~\cite{lee2025editsplat} & \xmark & \cmark & \xmark & Multi-view fusion in self-attn \\
% InterGSEdit~\cite{wen2025intergsedit} & User key view & \xmark & \cmark & 3D-consistent CA prior + fusion \\
% Edit360~\cite{huang2025edit360} & User anchor view & \xmark & \xmark & Video-diffusion $360^\circ$ propagation \\
% \midrule
% \textbf{Ours} & Attention-driven~\ref{sec:acp} & \cmark & \cmark & Joint SA + 3D CA align~\ref{sec:multi_attn_align} \\
% \bottomrule
% \end{tabular}
% }

% \end{table}

\begin{table}[t]
\centering
\footnotesize
\caption{Comparison of camera placement and attention alignment strategies in diffusion-based 3D editing. 
\textbf{CP}: editing-aware camera placement beyond the fixed COLMAP cameras used for reconstruction. 
\textbf{SA} / \textbf{CA}: self- / cross-attention alignment across views. 
Prior work performs at most one of SA or CA alignment, and only a few methods address camera placement at all---none of them in an attention-aware manner. 
\textbf{LB-Edit} is the first to jointly address all three.}
\label{tab:related_work_summary}
\setlength{\tabcolsep}{3pt}
\renewcommand{\arraystretch}{1.1}
\resizebox{\columnwidth}{!}{%
\begin{tabular}{@{}l c c c l@{}}
\toprule
\textbf{Method} & \textbf{CP} & \textbf{SA} & \textbf{CA} & \textbf{Consistency Strategy} \\
\midrule
IN2N~\cite{haque2023instructnerf2nerf}                 & \xmark               & \xmark & \xmark & Iterative dataset update \\
GaussianEditor~\cite{chen2024gaussianeditor}  & \xmark               & \xmark & \xmark & Semantic tracing + HGS \\
ViCA-NeRF~\cite{dong2023vicanerf}                 & Pixel-ratio views    & \xmark & \xmark & Depth-warped propagation \\
DGE~\cite{chen2024dge}                        & \xmark               & \cmark & \xmark & Ext.\ ST self-attn.\ + injection \\
GaussCtrl~\cite{wu2024gaussctrl}              & \xmark               & \cmark & \xmark & Cross-view latent attn. \\
VcEdit~\cite{wang2024vcedit}                  & \xmark               & \xmark & \cmark & 3D inverse splatting of CA \\
EditSplat~\cite{lee2025editsplat}             & \xmark               & \cmark & \xmark & Multi-view self-attn.\ fusion \\
InterGSEdit~\cite{wen2025intergsedit}         & User key view        & \xmark & \cmark & 3D CA prior + fusion \\
Edit360~\cite{huang2025edit360}               & User anchor view     & \xmark & \xmark & Video-diff.\ $360^{\circ}$ prop. \\
\midrule
\textbf{Ours (LB-Edit)} & \textbf{Attn.-driven} & \cmark & \cmark & \textbf{Joint SA + 3D CA align.} \\
\bottomrule
\end{tabular}%
}
\end{table}
% =========================
% Main paper (Sec. 3.1)
% =========================
\section{Method}
\label{sec:method}

\begin{figure*}[t]
    \centering
    \includegraphics[width=\textwidth]{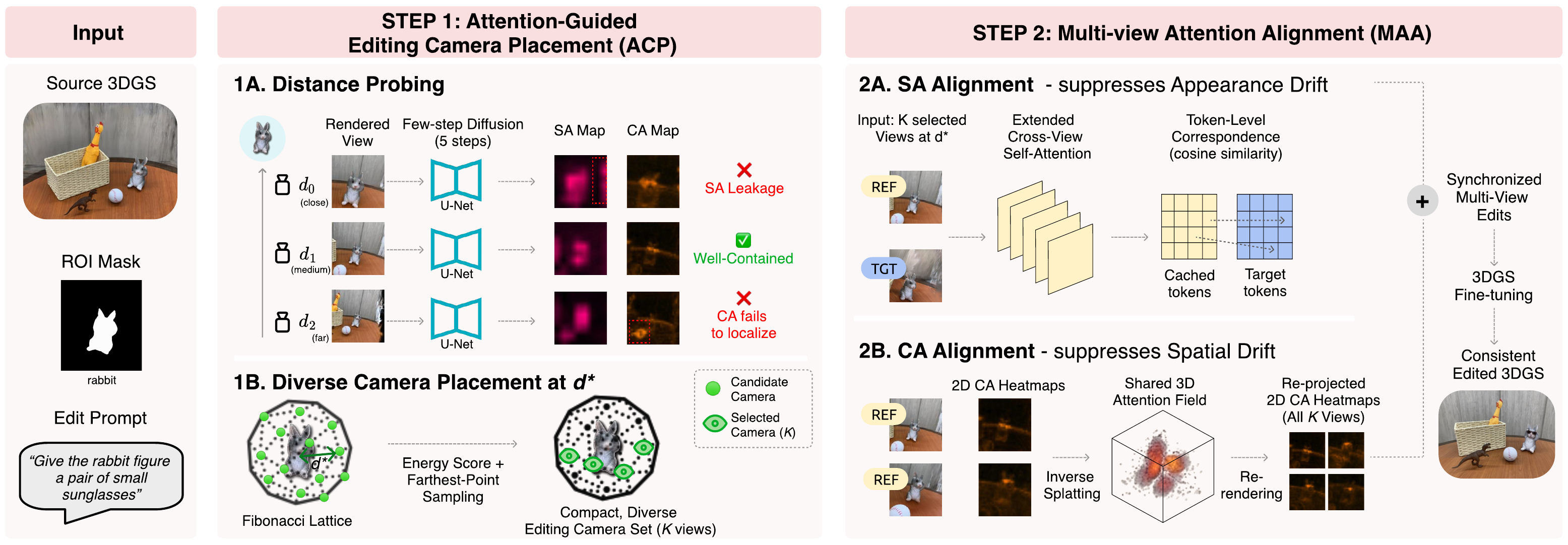}
    \caption{
    \textbf{Overview of the proposed framework.}
    Given a source 3DGS, an ROI mask, and a text editing prompt, our method first performs
    \textit{Attention-Guided Editing Camera Placement} (ACP) to identify editing-aware
    viewpoints. We probe candidate camera distances using few-step diffusion and select
    views where self-attention remains well contained within the target region while
    cross-attention reliably localizes the edit. From the optimal distance, we then sample
    a compact yet diverse set of editing cameras. Next, \textit{Multi-view Attention Alignment}
    (MAA) enforces consistency across the selected views: self-attention alignment reduces
    appearance drift through token-level correspondence, while cross-attention alignment
    suppresses spatial drift by aggregating and re-projecting a shared 3D attention field.
    The synchronized multi-view edits are finally used to fine-tune the 3DGS, producing a
    consistent edited 3D scene.
    }
    \Description{
    Overview of the proposed 3DGS editing pipeline. The method takes a source 3DGS, ROI mask,
    and edit prompt as input, selects attention-guided editing cameras, aligns self-attention
    and cross-attention across multiple views, and fine-tunes the 3DGS to obtain a consistent
    edited scene.
    }
    \label{fig:method_overview}
\end{figure*}

\subsection{Attention-Guided Editing Camera Placement}
\label{sec:acp}

Capturing the target object from suitable camera positions is a crucial first 
step for localized editing in complex scenes. We propose 
\textbf{Attention-Guided Editing Camera Placement (ACP)} to place a compact 
set of editing cameras that (i) yields stable, ROI-focused edits and 
(ii) covers the ROI geometry from diverse viewpoints. ACP operates in two 
stages: first choosing an attention-optimal camera distance where diffusion 
attention is well-contained within the ROI, then constructing a diverse and 
geometrically balanced editing camera set at that distance.

\paragraph{Problem formulation.}
Given a 3DGS scene $\mathcal{G}$ and a text edit instruction, we place 
$K$ editing cameras $\mathcal{C} = \{c_k\}_{k=1}^{K}$ such that applying 
a 2D diffusion editor to their rendered views produces ROI-focused edits with 
high 3D consistency. Let $\mathcal{M}_{\text{ROI}} \subseteq \mathcal{G}$ 
denote the ROI Gaussians obtained from 3D segmentation using language-guided 2D masks lifted into Gaussian space~\cite{langsam}, with center 
$\mathbf{c}$ and intrinsic scale 
$r_{\text{obj}} = \sqrt{\lambda_{\max}}$ from the largest eigenvalue of the 
opacity-weighted covariance. We define the canonical front direction 
$\mathbf{v}_{\text{front}} = 
\text{normalize}(\mathbf{c}_{\text{scene}} - \mathbf{c})$, where 
$\mathbf{c}_{\text{scene}}$ is the mean COLMAP camera position, representing 
the predominant viewing direction of the training cameras.

\paragraph{Distance probing via attention extraction.}
We probe candidate camera distances $d_i = m_i\, r_{\text{obj}}$ with 
multipliers $m_i$ sampled from a predefined range. For each $d_i$, we place 
a probe camera at $\mathbf{c} + d_i\,\mathbf{v}_{\text{front}}$, looking toward 
$\mathbf{c}$, and render a frontal view. We then run InstructPix2Pix for only 
5 denoising steps---sufficient because attention spatial structure stabilizes 
within the first few steps---to extract: (i) the projected 2D ROI mask $M_i$, 
(ii) the cross-attention map $A_i$ for edit tokens, and 
(iii) the self-attention matrix 
$\bar{S}_i \in \mathbb{R}^{N \times N}$, where $N$ is the number of spatial 
tokens at the lowest resolution.

\paragraph{Cross-attention concentration.}
Cross-attention determines where text tokens attend in the image, and thus 
where edits will be applied. A good editing camera distance should make 
cross-attention both \emph{localize on the ROI} and \emph{remain clean of 
background activation}. We capture these two properties as:
\begin{equation}
  S^{\text{ca}}_i = F_i \cdot (1 - \ell_i),
  \label{eq:s_ca}
\end{equation}
where $F_i \in [0, 1]$ is the F1 score between the thresholded 
cross-attention map $\hat{A}_i$ and the projected ROI mask $M_i$ 
(measuring ROI alignment), and $\ell_i \in [0, 1]$ is the 
90th-percentile cross-attention value in the background region 
(measuring residual background activation). Higher $S^{\text{ca}}_i$ 
indicates that the edit token correctly localizes the ROI without spilling 
onto the background.

\paragraph{Self-attention concentration.}
Self-attention governs how visual features propagate across spatial tokens. 
ROI feature leakage to the background---i.e., background tokens drawing 
information from ROI tokens---causes edits to bleed beyond the intended region. 
We measure this leakage as the average attention from background queries to 
ROI keys, normalized by ROI occupancy 
$o_i = \|\mathbf{m}\|_1 / N$ to enable fair comparison across camera distances 
with different ROI sizes:
\begin{equation}
  \mathcal{L}^{\text{sa}}_i = \frac{1}{o_i} \cdot 
  \frac{\bar{\mathbf{m}}^\top (\bar{S}_i\, \mathbf{m})}{\|\bar{\mathbf{m}}\|_1},
  \label{eq:sa_leak}
\end{equation}
where $\mathbf{m} \in \{0, 1\}^{N}$ is the downsampled ROI mask and 
$\bar{\mathbf{m}} = \mathbf{1} - \mathbf{m}$. Without the $1/o_i$ factor, 
larger ROIs would inflate the metric simply because background queries have 
more ROI keys to attend to. The corresponding concentration score is:
\begin{equation}
  S^{\text{sa}}_i = 1 - \mathcal{L}^{\text{sa}}_i,
  \label{eq:s_sa}
\end{equation}
with higher values indicating stronger containment of attention within the ROI.

\paragraph{Attention-optimal distance selection.}
The attention-optimal camera distance jointly maximizes both concentration scores:
\begin{equation}
  d^* = \arg\max_{d_i}\; S^{\text{ca}}_i + S^{\text{sa}}_i.
  \label{eq:dist_select}
\end{equation}
Both scores share comparable empirical ranges, allowing direct summation without 
explicit weighting. Camera distances with strong self-attention leakage or weak 
cross-attention localization are naturally penalized.

\paragraph{Energy-based camera placement.}
With $d^*$ fixed, we sample candidate camera directions on a sphere via 
Fibonacci lattice and place cameras at 
$\mathbf{p}_{\text{cam}} = \mathbf{c} + d^*\, \mathbf{d}$, looking toward 
$\mathbf{c}$. Each candidate camera is scored by:
\begin{equation}
  E_k = w_{\text{vis}}\, S^{\text{vis}}_k + w_{\text{can}}\, S^{\text{can}}_k,
  \label{eq:energy}
\end{equation}
where $S^{\text{vis}}_k$ is the projected ROI visibility ratio 
(higher when the ROI occupies more of the rendered view without occlusion), 
and $S^{\text{can}}_k$ favors canonical camera poses aligned with the intrinsic 
ROI axes. From the top-energy candidates, we select $K$ cameras via 
farthest-point sampling in angular space, ensuring diverse coverage without 
redundancy. Full definitions and implementation details are provided in 
Appendix~\ref{app:acp_details}.
\begin{figure*}[t]
    \centering
    \includegraphics[width=\textwidth]{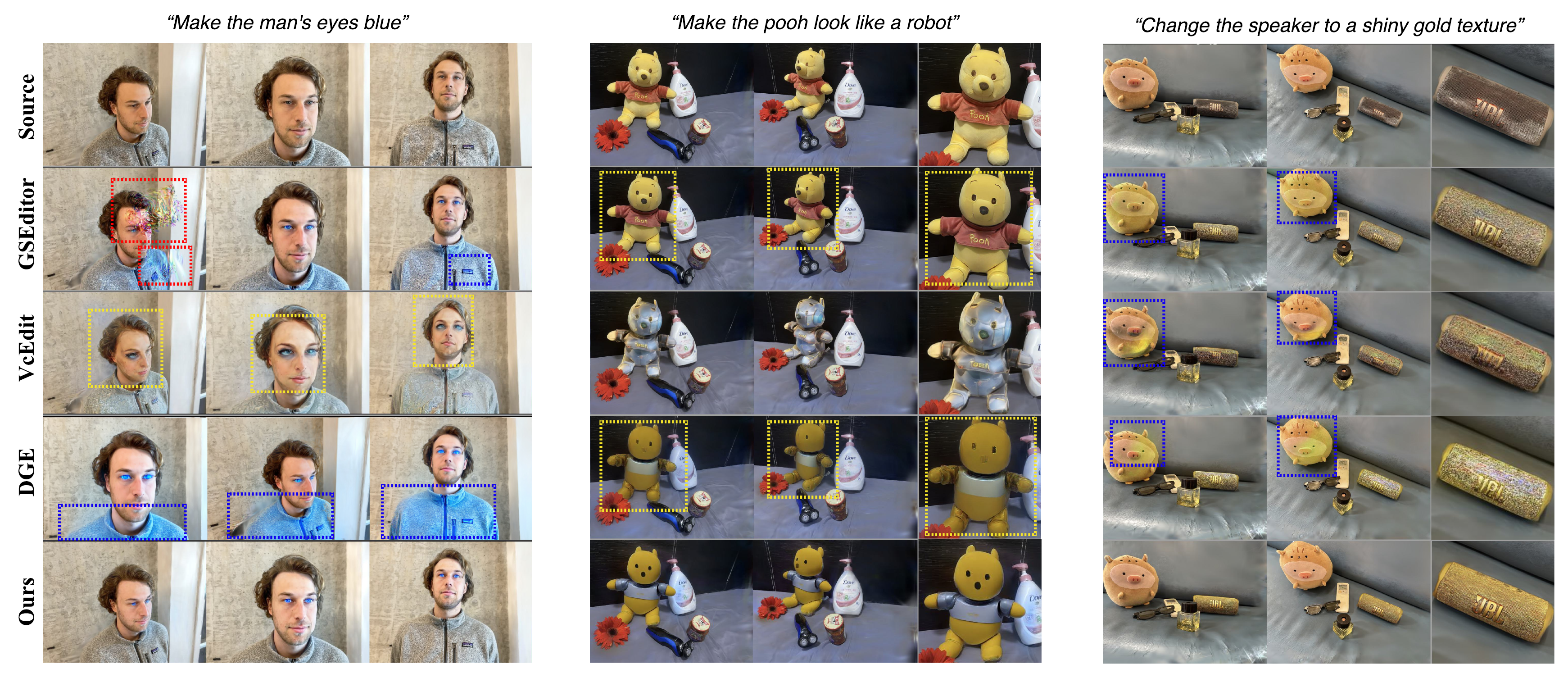}
    \caption{
    \textbf{Qualitative comparison.}
    Three editing tasks on multi-object and single-object scenes, compared
    against GSEditor, VcEdit, and DGE. Colored dashed boxes highlight
    characteristic baseline failures: red for view-dependent 3DGS artifacts
    and broken multi-view consistency, yellow for low instruction fidelity,
    and blue for edits that leak outside the target region. Our method
    produces edits that are simultaneously faithful to the instruction,
    localized to the ROI, and consistent across viewpoints.
    }
    \label{fig:qualitative_comparison}
\end{figure*}

\subsection{Multi-View Attention Alignment for Consistent Editing}
\label{sec:maa}

Even with carefully placed editing cameras, applying a 2D diffusion 
editor independently to each rendered view leads to two failure modes: 
\emph{appearance drift} (inconsistent style or shape across views) 
and \emph{spatial drift} (mis-aligned edit locations across views). 
In the U-Net architecture, self-attention governs how visual 
features propagate across spatial tokens and thus controls 
appearance coherence, while cross-attention determines the 
text-to-spatial mapping and thus controls where edits are applied.

Prior work addresses these failure modes individually: 
token-correspondence in self-attention harmonizes appearance 
across views~\cite{geyer2024tokenflow, chen2024dge}, while 
3D-lifted cross-attention enforces spatial 
consistency~\cite{wang2024vcedit}. We propose 
\textbf{Multi-View Attention Alignment (MAA)} that integrates 
both within a single U-Net forward pass, leveraging the compact 
editing camera set produced by ACP to make the joint mechanism efficient.

\paragraph{Reference--target framework.}
To keep alignment tractable, we avoid all-pairs attention exchange. 
At every diffusion timestep, we partition the rendered views from 
the $K$ editing cameras into a small set of \emph{reference views} 
$\mathcal{V}_{\text{ref}}$ and the remaining \emph{target views} 
$\mathcal{V}_{\text{tgt}}$. Reference views first perform a full 
U-Net forward pass to establish self- and cross-attention features; 
target views then inherit aligned features through the two replacements 
described below.

\paragraph{Self-attention alignment.}
We follow the token-correspondence paradigm of 
TokenFlow~\cite{geyer2024tokenflow} and DGE~\cite{chen2024dge}. 
During the reference pass, all reference views jointly perform 
\emph{Extended Cross-View Self-Attention}: keys and values are concatenated across reference views so each query token 
attends to tokens across the entire reference set. The resulting 
self-attention output is cached as 
$\mathbf{o}^{\text{self}}_{\text{ref}}$.

For each target view~$i$, we identify its nearest reference views 
by Euclidean distance between camera centers. To propagate features 
to a target token~$p$, we find the most similar reference token 
via cosine similarity on the L2-normalized, layer-normalized 
hidden states $\tilde{\mathbf{h}} = 
\text{LN}(\mathbf{h}) / \|\text{LN}(\mathbf{h})\|_2$:
\begin{equation}
  q^\star(p) = \arg\max_{q}\;
  \tilde{\mathbf{h}}_{i}(p)^\top\, \tilde{\mathbf{h}}_{\text{ref}}(q),
  \label{eq:sa_match}
\end{equation}
where $p, q$ index spatial tokens. Letting $\mathbf{o}^{\text{self}}_i$ 
denote the self-attention output of target view~$i$, we replace 
its value at token~$p$ with the matched reference output:
\begin{equation}
  \mathbf{o}^{\text{self}}_{i}(p) \leftarrow 
  \mathbf{o}^{\text{self}}_{\text{ref}}(q^\star(p)),
  \label{eq:sa_replace}
\end{equation}
followed by the standard residual connection. Each target token 
thus inherits the appearance representation already harmonized 
across reference views, suppressing appearance drift.

\paragraph{Cross-attention alignment.}
For spatial consistency, we lift cross-attention into a shared 
3D attention field via inverse splatting, following 
VcEdit~\cite{wang2024vcedit} and 
GaussianEditor~\cite{chen2024gaussianeditor}. From the reference 
pass we collect cross-attention probability maps 
$A^{(t)}_v \in \mathbb{R}^{HW \times |T|}$ at each denoising 
timestep~$t$, where $v \in \mathcal{V}_{\text{ref}}$, $H \times W$ 
is the attention map resolution, and $T$ denotes the edit-relevant 
tokens (all non-special tokens of the edit prompt). These 2D 
maps are lifted onto the ROI Gaussians as a per-Gaussian field 
$M^{(t)}_{\text{3D}}$ and re-rendered into each target view via 
differentiable Gaussian rasterization, yielding geometrically 
consistent maps $\hat{A}^{(t)}_i$ (full formulation in 
Appendix~\ref{app:inverse_splat}).

During the target U-Net pass, the standard cross-attention output 
$\text{softmax}(QK^\top/\sqrt{d})\,V$, with attention dimension 
$d$, is replaced by:
\begin{equation}
  \mathbf{o}^{\text{cross}}_{i} = \hat{A}^{(t)}_i\, \mathbf{V}_T,
  \label{eq:ca_replace}
\end{equation}
where $\mathbf{V}_T = W_V\, E_{\text{text}}[T] \in 
\mathbb{R}^{|T| \times d}$ are the value projections of 
edit-relevant tokens, $W_V$ is the value projection matrix, and 
$E_{\text{text}}$ is the text encoder embedding. Because all 
target views query the \emph{same} 3D attention field, the 
text-to-spatial mapping is view-consistent by construction, 
suppressing spatial drift.

\paragraph{Joint synchronization within a single forward pass.}
Both replacements are applied within the same U-Net forward pass: 
SA replacement (Eq.~\ref{eq:sa_replace}) at the \texttt{attn1} 
stage and CA replacement (Eq.~\ref{eq:ca_replace}) at the 
\texttt{attn2} stage of each transformer block. While each 
mechanism has appeared individually in prior 
work~\cite{geyer2024tokenflow, chen2024dge, wang2024vcedit}, to 
our knowledge ours is the first 3D editing pipeline that 
synchronizes both within a single 2D editing pass. Combined with 
the compact editing camera set produced by ACP, this enables 
high-quality multi-view-consistent editing on as few as 5 editing 
cameras.

% Put this near the end of the Experiments/Evaluation section,
% before \section{Ablation Studies}.
\begin{figure*}[t]
    \centering
    \begin{subfigure}{0.53\textwidth}
        \centering
        \includegraphics[
            width=\linewidth,
            clip,
            trim=7cm 0cm 7cm 0cm
        ]{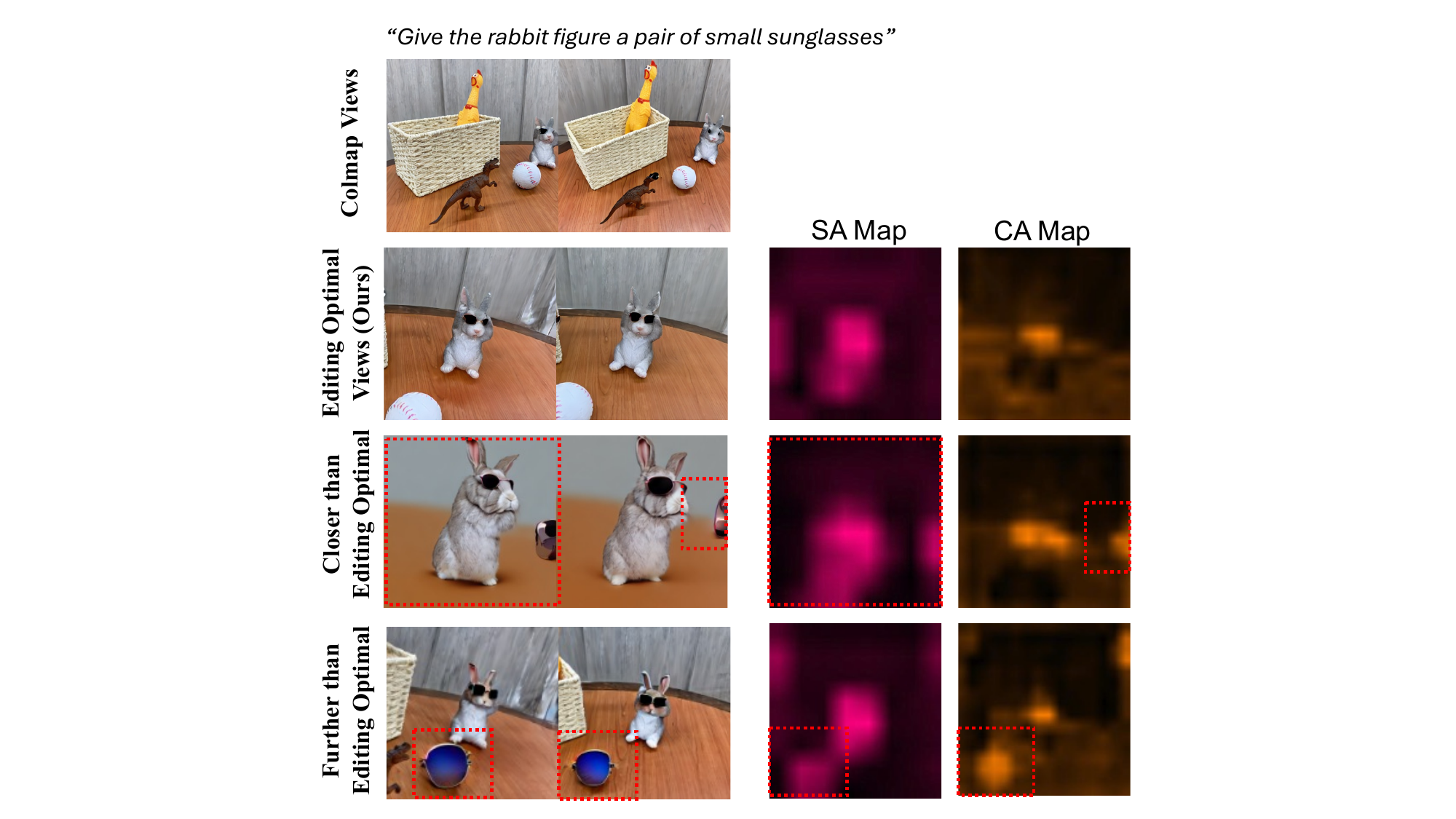}
        \caption{}
        \label{fig:ablation-a}
    \end{subfigure}
    \hspace{0.005\textwidth}
    \begin{subfigure}{0.445\textwidth}
        \centering
        \includegraphics[
            width=\linewidth,
            clip,
            trim=9cm 0cm 8cm 0cm
        ]{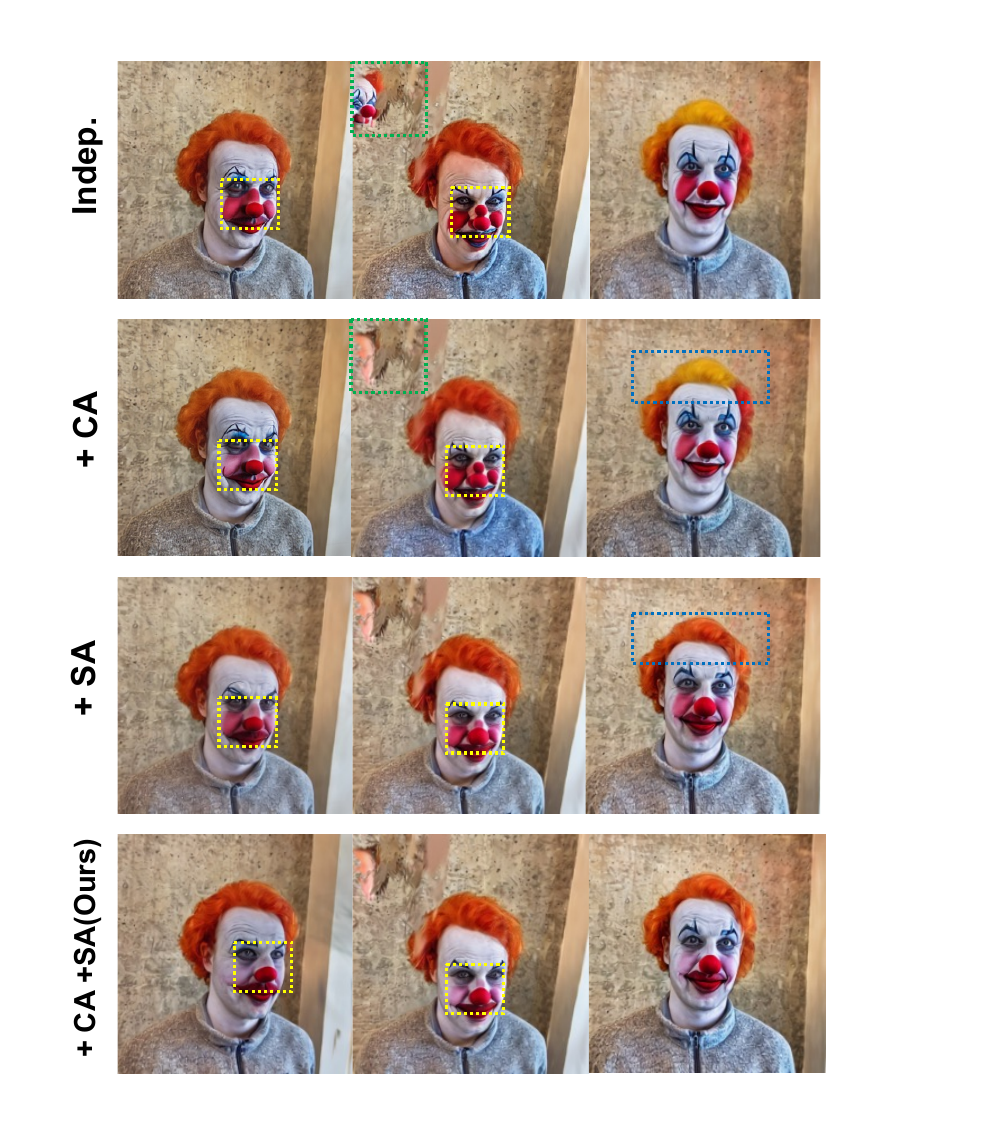}
        \caption{}
        \label{fig:ablation-b}
    \end{subfigure}

    \caption{
    \textbf{Ablation studies.}
    (a) Ablation on Attention-Guided Editing Camera Placement
    (Sec.~\ref{sec:acp}).
    (b) Ablation on Multi-View Attention Alignment for consistent editing
    (Sec.~\ref{sec:maa}).
    }
    \label{fig:ablation}
\end{figure*}

\section{Experiments}
\label{sec:experiments}

We evaluate LB-Edit on scenes represented by vanilla 3D Gaussian
Splatting. Given a trained 3DGS model, our pipeline
proceeds in three stages: (i) ACP selects a compact set of editing cameras at
the attention-optimal distance (Sec.~\ref{sec:acp}); (ii) MAA edits the
rendered views with InstructPix2Pix~\cite{brooks2023instructpix2pix} while
jointly aligning self- and cross-attention across views (Sec.~\ref{sec:maa});
and (iii) the 3DGS model is fine-tuned on the synchronized edits.

Depending on scene complexity, we use between $5$ and $20$ editing cameras,
with as few as $5$ on the simpler scenes. Fine-tuning follows the loss
schedule of IN2N~\cite{haque2023instructnerf2nerf}, combining an $L_1$ term
with an LPIPS term. To restrict optimization to the target object, we lift a
2D segmentation mask onto the Gaussians via inverse splatting
(Appendix~\ref{app:inverse_splat}) and update only those Gaussians
belonging to the resulting 3D ROI mask $\mathcal{M}_{\text{ROI}}$. All
experiments are run on a single NVIDIA RTX A6000 GPU.

\subsection{Experimental Setup}

\paragraph{Baselines.}
We compare against three representative diffusion-based 3DGS editors:
GaussianEditor~\cite{chen2024gaussianeditor}, VcEdit~\cite{wang_vcedit_2024},
and DGE~\cite{chen2024dge}. All three rely on COLMAP training cameras for
editing supervision; they differ in how they fuse per-view edits back into
the 3D representation, using SDS-based optimization (GaussianEditor) or
iterative reconstruction from edited image sets (VcEdit, DGE). None of them
performs editing-aware camera placement, and each aligns at most one of
self- or cross-attention across views (see
Table~\ref{tab:related_work_summary}).

\paragraph{Datasets.}
We evaluate on five scenes drawn from two standard benchmarks. From
3D-OVS we use \textit{room}, \textit{covered\_desk}, and
\textit{blue\_sofa}, all forward-facing scenes containing multiple distinct
objects---the regime in which editing-aware camera placement matters most.
From IN2N~\cite{haque2023instructnerf2nerf} we use \textit{face}
(forward-facing) and \textit{bear} (forward-facing 360$^\circ$). We
construct 20 text editing prompts per scene; the full list, together with
the segmentation prompts used to obtain $\mathcal{M}_{\text{ROI}}$ and the
source/target prompts used for CLIP similarity, is provided in Table~6 of
the appendix.

\paragraph{Evaluation metrics.}
We report CLIP directional similarity
(CLIPsim)~\cite{patashnik2021styleclip} to measure how well the edited
scene reflects the textual instruction. Because CLIP-based scores correlate
imperfectly with perceptual editing quality---particularly when an edit
affects the global appearance of an image rather than the intended object,
a failure mode we examine in Sec.~\ref{sec:ablation}---we additionally
conduct a user study covering instruction fidelity, multi-view consistency,
and editing locality. Finally, we report wall-clock latency to quantify the
computational benefit of editing with a compact view set.

\subsection{Qualitative Comparisons}

Figure~\ref{fig:qualitative_comparison} compares our edits with the
baselines on three representative tasks. Two patterns recur across the
baselines. First, on scenes that contain multiple objects, the ROI often
occupies only a small fraction of the COLMAP renderings, and the 2D editor
either fails to apply the requested edit or applies it weakly (yellow
boxes). Because ACP places editing cameras at the distance where the
diffusion model's attention is best contained within the ROI, our
renderings present the target object at a scale that the editor can act on
reliably.

Second, even when a baseline does produce a recognizable edit, the per-view
results disagree: appearance drifts in color or texture, and the spatial
extent of the edit shifts between views (red and blue boxes). DGE
partially mitigates this through extended self-attention but leaves
cross-attention unconstrained, while VcEdit aligns cross-attention but not
self-attention. MAA synchronizes both within a single forward pass, and the
resulting fused 3DGS is correspondingly more stable across viewpoints.

\subsection{Quantitative Comparisons}

\begin{table}[h]
\centering
\caption{CLIP directional similarity ($\uparrow$) across scenes from the
3D-OVS and IN2N datasets. Best per row in bold.}
\label{tab:overall-quant}
\begin{tabular}{lcccc}
\toprule
\textbf{Scene} & \textbf{GSEditor} & \textbf{VcEdit} & \textbf{DGE} & \textbf{Ours} \\
\midrule
\multicolumn{5}{l}{\textit{3D-OVS}} \\
blue\_sofa     & 9.70  & 7.54  & 11.79          & \textbf{13.10} \\
covered\_desk  & 4.72  & 16.22 & \textbf{19.84} & 18.48 \\
room           & 7.21  & 12.79 & 19.46          & \textbf{19.78} \\
\midrule
\multicolumn{5}{l}{\textit{IN2N}} \\
face           & 9.50  & 10.10 & 19.30          & \textbf{23.00} \\
bear           & 12.00 & \textbf{23.80} & 19.20 & 16.20 \\
\bottomrule
\end{tabular}
\end{table}

\begin{table}[t]
\centering
\small
\setlength{\tabcolsep}{6pt}
\begin{tabular}{lcccc}
\toprule
Criterion & Ours & GSEditor & VcEdit & DGE \\
\midrule
Instruction fidelity   & \textbf{41.7\%} & 19.8\% & 18.7\% & 19.8\% \\
Multi-view consistency & \textbf{37.7\%} & 22.2\% & 22.6\% & 17.5\% \\
Editing locality       & \textbf{32.9\%} & 30.6\% & 21.4\% & 15.1\% \\
\midrule
Overall                & \textbf{37.4\%} & 24.2\% & 20.9\% & 17.5\% \\
\bottomrule
\end{tabular}
\caption{User study results (21 participants, 14 tasks). Each entry reports
the percentage of trials on which a method was preferred under the given
criterion.}
\label{tab:userstudy_preferences}
\end{table}

\begin{figure}[t]
    \centering
    \includegraphics[
        width=\columnwidth,
        clip,
        trim=1.0cm 0.7cm 1.0cm 0.6cm
    ]{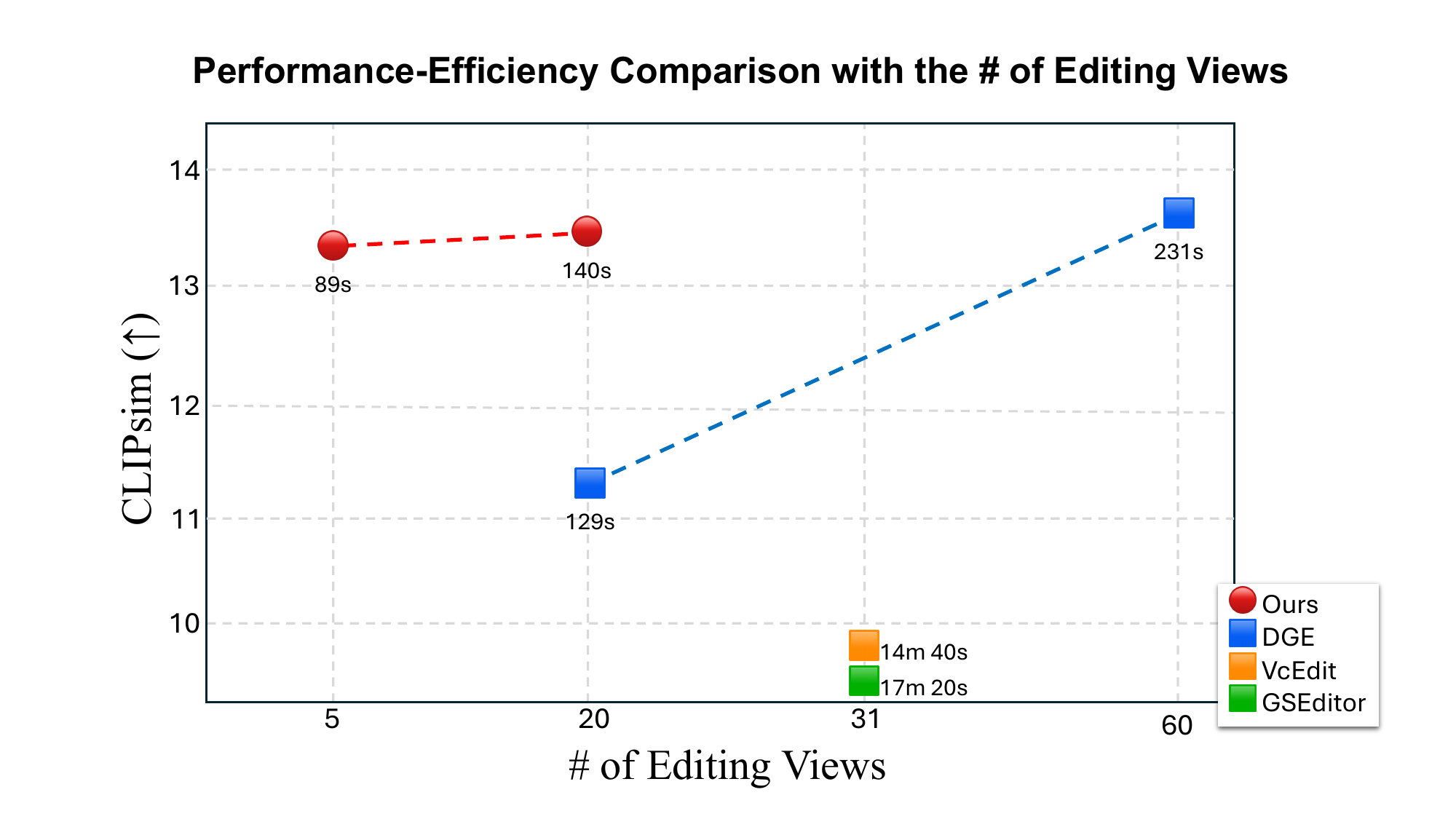}
    \caption{
    \textbf{Performance--efficiency trade-off on the \textit{room} scene
    (``Turn the dinosaur color into green'').}
    CLIP similarity ($\uparrow$) against the number of editing views.
    With only 5 ACP-selected editing cameras (89\,s), our method matches the
    CLIPsim of DGE at 60 views (231\,s) and exceeds that of VcEdit and
    GSEditor at 20 views ($>$14\,min), yielding up to a $7\times$ reduction
    in editing latency.
    }
    \label{fig:graph}
\end{figure}

Table~\ref{tab:overall-quant} reports CLIPsim across all five scenes. Our
method achieves the highest score on four of the five, indicating that the
edits are better aligned with the textual instruction overall. The one
exception is \textit{bear}, where VcEdit obtains a higher CLIPsim; we
analyze this case Ablation and find that CLIPsim can
reward edits that bleed color across the entire image---a pathology that
ACP explicitly avoids.

To assess perceptual quality, we conduct a user study with 21 participants
across 14 editing tasks. Participants rate edits along three axes:
\emph{instruction fidelity}---whether the edit follows the text prompt;
\emph{multi-view consistency}---whether the edited scene is coherent across
viewpoints; and \emph{editing locality}---whether the edit remains confined
to the intended region. As shown in
Table~\ref{tab:userstudy_preferences}, our method is preferred on every
criterion, with the largest margin on instruction fidelity, where ACP's
ROI-focused viewpoint selection has the most direct effect.

Figure~\ref{fig:graph} reports the trade-off between editing quality and
computation on the \textit{room} scene with the prompt ``Turn the dinosaur
color into green.'' Because ACP returns a compact, geometrically diverse
view set, our method matches or exceeds baseline CLIPsim while using only
$5$--$20$ editing cameras depending on the scene, with as few as $5$ on the
simpler ones---reducing latency by up to $7\times$ relative to GSEditor and
VcEdit.

\section{Ablation Studies}
\label{sec:ablation}
\paragraph{Ablation on editing-view selection.}
Figure~\ref{fig:ablation-a} compares editing results under different view distances.
\textbf{COLMAP Views} often fail to produce the intended edit because the ROI occupies only a small portion of the rendered image, resulting in weak and noisy diffusion guidance.
\textbf{Editing-Optimal Views (Ours)} successfully apply the edit while keeping both self-attention (SA) and cross-attention (CA) concentrated on the ROI.
\textbf{Closer-than-Optimal} views cause drastic global changes; these cases are primarily filtered out by high SA leakage scores, indicating that the editing signal propagates from the ROI to non-ROI regions.
\textbf{Further-than-Optimal} views may edit the ROI but also introduce unintended edits in the background (for example, the baseball becoming part of the sunglasses); such cases are rejected due to low CA-to-mask alignment (F1) and elevated SA leakage.

\paragraph{Ablation on multi-view attention alignment.}
Figure~\ref{fig:ablation-b} presents the impact of aligning SA and CA across views.
\textbf{Indep.} applies the 2D editor independently per view, leading to semantic and geometric inconsistencies across viewpoints.
\textbf{+CA} reduces spatial drift by sharing cross-attention in 3D space, suppressing spurious activations that appear in intermediate views.
\textbf{+SA} improves appearance consistency by harmonizing semantics such as color, texture, and facial attributes across views.
Finally, \textbf{+CA+SA} combines both benefits, yielding the most consistent multi-view results and notably improving the stability of fine-grained internal structures (such as the nose), which are particularly prone to view-dependent inconsistencies.
\section{Conclusion}
\label{sec:conclusion}

We presented \textbf{LB-Edit}, a framework for localized and consistent text-driven 3D Gaussian Splatting editing.
Our key insight is that camera placement and multi-view consistency should be treated as attention-aware problems: by analyzing how diffusion self- and cross-attention interact with the region of interest, we can both select where to look and ensure that edits agree across viewpoints.

Attention-Guided Editing Camera Placement (ACP) probes a small
set of candidate distances to identify where diffusion attention is
well-localized to the ROI, then constructs a compact, geometrically
diverse camera set at that distance.
Multi-View Attention Alignment jointly synchronizes self-attention features and cross-attention maps across views through token-level correspondence and a shared 3D attention field, suppressing both appearance and spatial drift.
Together, these components enable high-quality localized edits with as few as 5 views, achieving the highest user preference across all evaluation criteria while reducing editing latency by up to $7\times$ compared to methods that rely on 20--60 views.

\paragraph{Limitations and future work.}
Our method inherits the generation quality ceiling of the underlying 2D diffusion editor; edits that InstructPix2Pix cannot produce in 2D will not succeed in 3D either.
The current distance probing assumes a single dominant ROI per edit instruction; extending AGEVS to handle multiple disjoint regions simultaneously is an interesting direction.
Finally, integrating more expressive diffusion backbones or video diffusion models could further improve both the fidelity and temporal coherence of 3D edits.

% ---------------------------------------------------------------
% Bibliography
% acmart usually uses ACM-Reference-Format.

\bibliographystyle{ACM-Reference-Format}
% \bibliography{final}
\bibliography{references}

\begin{figure*}[!p]
    \centering
    \includegraphics[width=\linewidth,height=0.9\textheight,keepaspectratio]{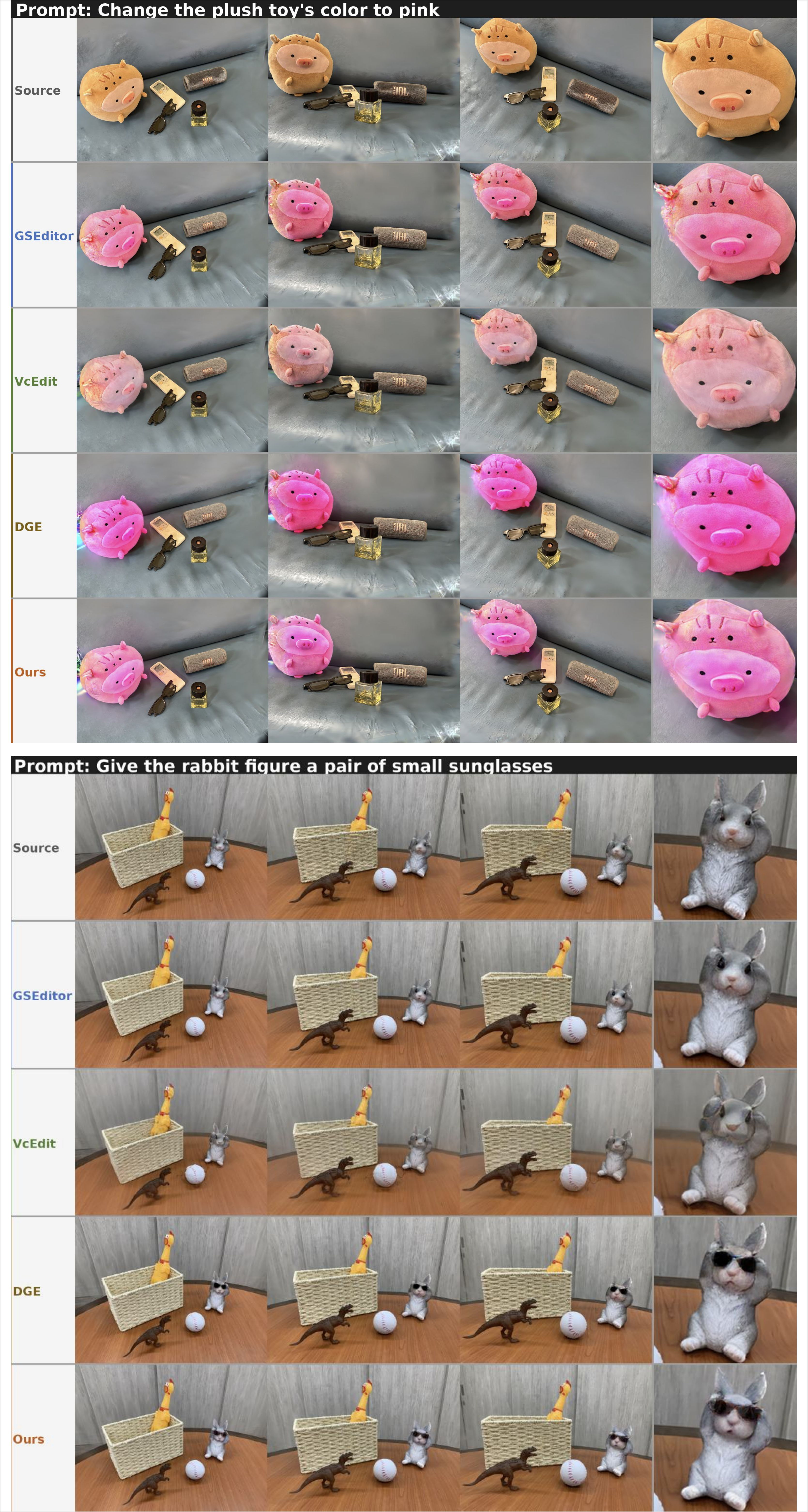}
    \caption{%
    We provide additional qualitative comparisons. Overall, our method consistently achieves higher instruction fidelity, better multiview consistency, and more precise editing locality compared to the baselines.
    }
    \label{fig:sup-1}
\end{figure*}

\begin{figure*}[!p]
    \centering
    \includegraphics[width=\linewidth,height=0.9\textheight,keepaspectratio]{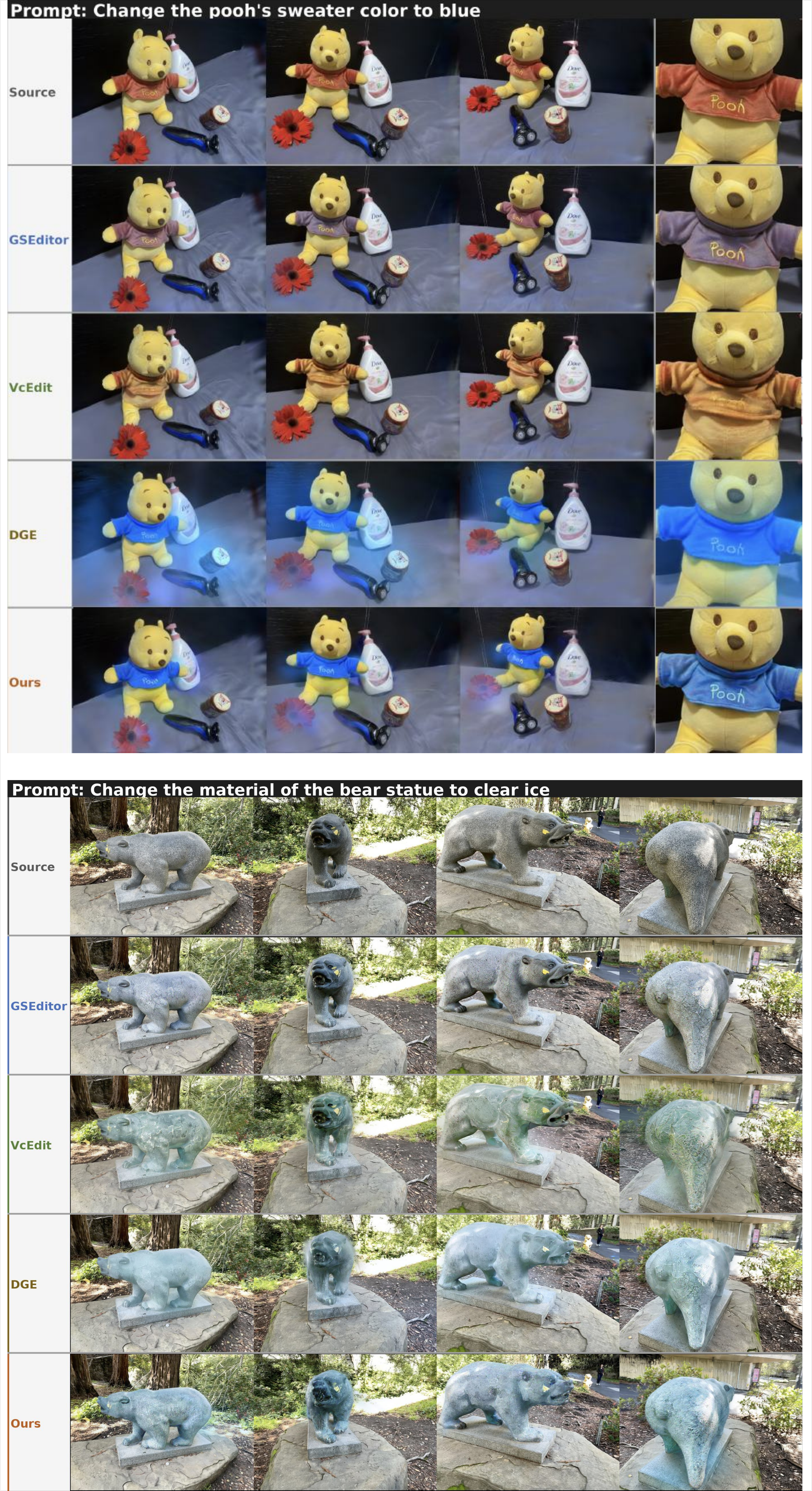}
    \caption{%
    Additional qualitative comparisons.
    }
    \label{fig:sup-2}
\end{figure*}
% ---------------------------------------------------------------
% Appendices
% For SIGGRAPH, check whether supplementary material should be separate.
% If the template/project allows appendix in main PDF, keep these.
% Otherwise, move them to a supplementary file.

\clearpage
\appendix

\appendix

\section{ACP Details}
\label{app:acp_details}

\subsection{Attention Statistics Stabilize Early}
ACP uses short diffusion runs solely to estimate attention statistics, not to 
produce final edited images. Empirically, both cross-attention 
(token-to-spatial localization) and self-attention (token-to-token propagation) 
exhibit a stable coarse spatial layout within the early denoising stages, even 
before the final appearance has converged. We therefore run InstructPix2Pix for 
only 5 steps when extracting attention maps, amortizing the camera-placement 
cost. Full editing still uses the standard denoising schedule.

\subsection{Distance Probing Setup}
For each candidate camera distance $d_i = m_i\, r_{\text{obj}}$ 
(with $m_i \in \{5.0, 6.0, 7.0, 8.0\}$, adjusted per scene), we place a probe 
camera at $\mathbf{c} + d_i\, \mathbf{v}_{\text{front}}$, looking toward 
$\mathbf{c}$, and render a frontal view. A single InstructPix2Pix forward pass 
collects:
\begin{enumerate}
  \item[(i)] the projected 2D ROI mask 
  $M_i \in \{0, 1\}^{H \times W}$ obtained by rendering the 3D ROI Gaussians 
  $\mathcal{M}_{\text{ROI}}$ into the probe view;
  \item[(ii)] the cross-attention map 
  $A_i \in [0, 1]^{H' \times W'}$, averaged over heads and timesteps for 
  edit-relevant text tokens, taken at the $16 \times 16$ resolution;
  \item[(iii)] the self-attention matrix 
  $\bar{S}_i \in \mathbb{R}^{N \times N}$, averaged over heads and timesteps 
  at the lowest resolution ($N = 8 \times 8 = 64$ tokens), chosen for both 
  robustness and computational efficiency.
\end{enumerate}
When multiple candidate camera distances are probed, they are processed in a 
single batched forward pass to reduce wall time.

\subsection{Cross-Attention Concentration: Detailed Definitions}
\label{app:acp_ca_details}

The CA concentration score in Eq.~\ref{eq:s_ca} combines two complementary 
measurements: alignment with the ROI ($F_i$) and absence of background 
activation ($\ell_i$).

\paragraph{ROI alignment via F1.}
We binarize the cross-attention map $A_i$ at its 75th percentile to obtain the 
high-activation region $\hat{A}_i$, then compute precision, recall, and F1 
against the projected ROI mask $M_i$:
\begin{equation}
  P_i = \frac{\sum \hat{A}_i \odot M_i}{\sum \hat{A}_i + \epsilon}, 
  \quad
  R_i = \frac{\sum \hat{A}_i \odot M_i}{\sum M_i + \epsilon}, 
  \quad
  F_i = \frac{2\, P_i\, R_i}{P_i + R_i + \epsilon},
  \label{eq:ca_f1}
\end{equation}
where $\odot$ denotes element-wise multiplication and $\epsilon$ prevents 
division by zero. The 75th percentile threshold balances sensitivity to weak 
activations with robustness to noise; we verified that thresholds in $[70, 80]$ 
yield similar placement outcomes.

\paragraph{Background activation via 90th percentile.}
We define $\ell_i$ as the 90th-percentile cross-attention value within the 
background region:
\begin{equation}
  \ell_i = Q_{0.9}\bigl( A_i[\bar{M}_i > 0.5] \bigr),
  \label{eq:ca_bg_leak}
\end{equation}
where $\bar{M}_i = 1 - M_i$ selects background pixels. We use the 90th percentile 
rather than the mean (insensitive to localized hot spots) or the maximum 
(overly noise-sensitive); it captures the strongest persistent background 
activation while remaining robust to outliers.

\subsection{Self-Attention Concentration: Detailed Definitions}
\label{app:acp_sa_details}

\paragraph{Why background-to-ROI attention measures leakage.}
In the U-Net self-attention layer, the output for a background query $q$ is 
$\sum_k \bar{S}_i[q, k]\, V[k]$, where $V$ are value vectors. When background 
queries place strong attention on ROI keys, ROI value vectors mix into 
background outputs, propagating ROI features to the background. The numerator 
of Eq.~\ref{eq:sa_leak} aggregates exactly this background-to-ROI attention mass.

\paragraph{Why occupancy normalization.}
The raw quantity 
$\bar{\mathbf{m}}^\top (\bar{S}_i \mathbf{m}) / \|\bar{\mathbf{m}}\|_1$ 
is biased by ROI size: as the ROI grows, background queries naturally have more 
ROI keys to attend to, inflating the metric even without genuine attention bias. 
To make a hypothetical example concrete, consider a uniform-attention setting 
where $\bar{S}_i[q, k] = 1/N$ for all $q, k$. Substituting into 
Eq.~\ref{eq:sa_leak}'s numerator yields exactly $o_i$, so 
$\mathcal{L}^{\text{sa}}_i = 1$ regardless of attention pattern. This confirms 
that without normalization, the metric is dominated by ROI size rather than the 
genuine spatial structure of attention. Dividing by $o_i$ removes this size 
dependency, making $\mathcal{L}^{\text{sa}}_i$ a scale-invariant measure of 
leakage intensity.

\paragraph{Resolution choice.}
We compute $\mathcal{L}^{\text{sa}}_i$ at the lowest U-Net self-attention 
resolution ($8 \times 8$, $N = 64$) for two reasons: 
(i) low-resolution attention captures global propagation patterns rather than 
local texture details, and (ii) the smaller matrix size keeps the ACP overhead 
negligible.

\subsection{Attention-Optimal Distance Selection}
The attention-optimal camera distance is selected by maximizing the joint 
concentration score (Eq.~\ref{eq:dist_select}). Both $S^{\text{ca}}_i$ and 
$S^{\text{sa}}_i$ are bounded in approximately $[0, 1]$ in our experiments, 
allowing direct summation. We empirically verified that the selected $d^*$ is 
stable across all 5 tested scenes.

\subsection{Energy-Based Camera Placement and Diversity}

With $d^*$ fixed, we sample candidate camera directions $\mathbf{d}$ on the 
unit sphere $\mathbb{S}^2$ via Fibonacci lattice and place cameras at 
$\mathbf{p}_{\text{cam}} = \mathbf{c} + d^*\, \mathbf{d}$, looking toward 
$\mathbf{c}$.

\paragraph{Forward-facing scenes.}
For forward-facing scenes (3D-OVS: \textit{room}, 
\textit{covered\_desk}, \textit{blue\_sofa}), candidate cameras are restricted 
to a cone of half-angle $60^\circ$ centered on $\mathbf{v}_{\text{front}}$, 
ensuring cameras remain on the observable side of the scene.

\paragraph{$360^\circ$ scenes.}
For $360^\circ$ scenes (textit{bear}), the full sphere is 
used. Since PCA-derived $\mathbf{v}_{\text{front}}$ is ambiguous on a full sphere, 
we instead set $\mathbf{v}_{\text{front}}$ to the direction from the ROI center 
toward a randomly chosen COLMAP camera. Candidate cameras are filtered to an 
elevation band of $\pm 20^\circ$ around the equatorial plane, restricting them 
to roughly horizontal orbits and avoiding degenerate top-down or bottom-up 
viewpoints. The canonical alignment weight $w_{\text{can}}$ is set to $0$ so all 
camera directions are scored equally by visibility.

\paragraph{Energy score components.}
Each candidate camera $k$ is scored by Eq.~\ref{eq:energy}, with:
\begin{itemize}
  \item $S^{\text{vis}}_k \in [0, 1]$: the fraction of ROI Gaussians visible 
  (non-occluded) when projected into camera $k$.
  \item $S^{\text{can}}_k \in [0, 1]$: alignment between the viewing direction 
  $\mathbf{v}_{\text{view}} = 
  \text{normalize}(\mathbf{c} - \mathbf{p}_{\text{cam}})$ and the intrinsic ROI 
  axes $\mathbf{v}_{\text{front}}$ (primary) and $\mathbf{v}_2$ (secondary), 
  estimated from weighted PCA on the ROI Gaussians:
  \begin{equation}
    S^{\text{can}}_k = \max\!\left(
      |\mathbf{v}_{\text{view}} \cdot \mathbf{v}_{\text{front}}|,\; 
      0.8\, |\mathbf{v}_{\text{view}} \cdot \mathbf{v}_2|
    \right).
    \label{eq:s_can}
  \end{equation}
  The factor $0.8$ slightly down-weights the secondary axis to prefer the 
  primary canonical orientation. Weights $w_{\text{vis}} = 1.0$ and 
  $w_{\text{can}} = 0.5$ are used for forward-facing scenes; 
  $w_{\text{can}} = 0$ for $360^\circ$ scenes as noted above.
\end{itemize}

\paragraph{Diversity-aware camera selection.}
From the top-ranked candidate cameras (top 20\% for forward-facing scenes, all 
candidates for $360^\circ$ scenes), we select $K$ cameras by farthest-point 
sampling in angular space. Each new camera maximizes a combination of its 
minimum angular distance to the already-selected set and its normalized energy:
\begin{equation}
  \arg\max_{k\, \in\, \text{cand}} \left( 
    \min_{j\, \in\, \text{sel}} \angle(\mathbf{d}_k, \mathbf{d}_j) 
    + w_\phi\, \Delta\phi_{\min} 
    + w_y\, \Delta y_{\min} 
    + \bar{E}_k 
  \right),
  \label{eq:fps_selection}
\end{equation}
where $\Delta\phi_{\min}$ and $\Delta y_{\min}$ are the minimum azimuth and 
elevation differences to the selected set, $w_\phi$ and $w_y$ are diversity 
weights, and $\bar{E}_k$ is the energy score normalized to $[0, 1]$. The final 
editing camera set is sorted by azimuth around $\mathbf{c}$ for stable ordering.

\subsection{Implementation Notes}
All ACP computations are negligible compared to full multi-view diffusion 
editing and are amortized across subsequent 3DGS optimization. Cross-attention 
maps are aggregated over edit-relevant tokens at $16 \times 16$ resolution and 
averaged over the 5 probing timesteps. When multiple candidate camera distances 
are probed, they are processed in a single batched IP2P forward pass 
(batch size equal to the number of candidates) to reduce wall time.

\begin{figure*}[t]
    \centering
    \includegraphics[
        width=0.97\textwidth,
        trim=0cm 13cm 0cm 0cm,
        clip
    ]{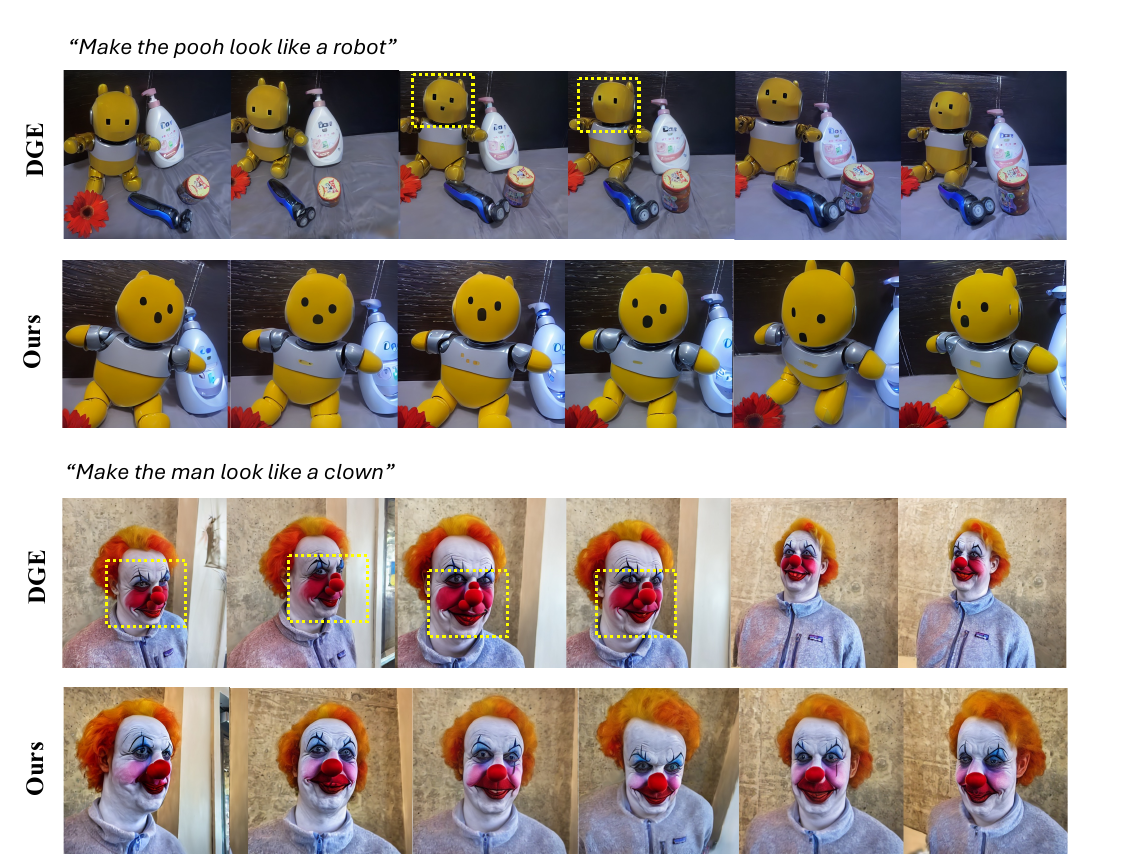}
    \caption{\textbf{Consistent multi-view edits.} DGE (COLMAP cameras) vs.\ Ours (ACP cameras) on two instance-level transformations. DGE exhibits view-dependent artifacts such as missing or duplicated geometry (yellow boxes), while our method produces consistent edits across all viewpoints.}
    \label{fig:app-multiview_edits}
\end{figure*}

\section{Multi-view Attention Alignment Details}
\label{app:maa_details}

\subsection{Reference--Target Partitioning}
\label{app:maa_ref_target}

For a set of $K$ editing views, we partition them into reference 
and target views with a fixed reference count 
$|\mathcal{V}_{\text{ref}}| = \lfloor K / B \rfloor$, where $B = 5$ 
is the batch size used in the implementation. This corresponds to 
roughly $20\%$ of views serving as references at each timestep:
\begin{itemize}
  \item $K = 5$: $|\mathcal{V}_{\text{ref}}| = 1$
  \item $K = 20$: $|\mathcal{V}_{\text{ref}}| = 4$
\end{itemize}
The reference set is re-sampled at every denoising timestep, so 
any given view may serve as a reference at one timestep and as a 
target at another. This stochastic role-swapping further diversifies 
the propagated features over the denoising trajectory.

\begin{table*}[t]
\caption{\textbf{Runtime breakdown of different editing methods} on the \textit{room} scene. $^\dagger$CLIPsim outlier caused by global color bleeding (see text).}
\label{tab:runtime_breakdown}
\centering
\small
\setlength{\tabcolsep}{4pt}
\renewcommand{\arraystretch}{1.15}
\begin{tabular}{@{}
>{\raggedright\arraybackslash}p{2.2cm}
>{\centering\arraybackslash}p{1.6cm}
>{\centering\arraybackslash}p{2.3cm}
>{\centering\arraybackslash}p{2.5cm}
>{\centering\arraybackslash}p{2.1cm}
>{\centering\arraybackslash}p{1.9cm}
>{\centering\arraybackslash}p{1.3cm}
@{}}
\toprule
\textbf{Method (\# Editing Views)} &
\textbf{Total Time} &
\textbf{Editing View Generation} &
\textbf{Multi-view 2D Image Editing} &
\textbf{3DGS Fine-tuning} &
\textbf{3DGS Masking} &
\textbf{CLIPsim} \\
\midrule
GSEditor (31) & 17m44s & -- & 14m40s & 3m04s & -- & 12.71 \\
VcEdit (31)   & 20m31s & -- & 16m10s & 4m21s & -- & 8.96 \\
DGE (20) & 137s & -- & 60s & 68s & 9s & 11.50 \\
DGE (5)  & 84s  & -- & 14s & 67s & 3s & \textit{15.30}$^\dagger$ \\
Ours (20) & 139s & 5s & 39s & 86s & 9s & 13.70 \\
Ours (5)  & 88s  & 6s & 7s & 70s & 5s & 13.40 \\
\bottomrule
\end{tabular}
\end{table*}

\subsection{Self-Attention Alignment: Detailed Operations}
\label{app:maa_sa_align}

\paragraph{Extended spatio-temporal attention at the reference pass.}
At each \texttt{attn1} layer, the reference forward pass concatenates 
keys and values from all reference views along the spatial dimension. 
Specifically, given per-view query, key, value tensors 
$Q_v, K_v, V_v \in \mathbb{R}^{HW \times d}$ for 
$v \in \mathcal{V}_{\text{ref}}$, we compute:
\begin{equation}
  K_{\text{ext}} = [K_{v_1}; K_{v_2}; \dots] 
  \in \mathbb{R}^{|\mathcal{V}_{\text{ref}}| HW \times d},
  \quad
  V_{\text{ext}} = [V_{v_1}; V_{v_2}; \dots],
  \label{eq:ext_kv}
\end{equation}
and the self-attention output for reference view $v$ becomes:
\begin{equation}
  \mathbf{o}^{\text{self}}_{v} = \text{softmax}\!\left(
    \frac{Q_v\, K_{\text{ext}}^\top}{\sqrt{d}}
  \right) V_{\text{ext}}.
  \label{eq:ref_sa}
\end{equation}
This output is cached and reused during target view processing.

\paragraph{Hidden state caching for similarity matching.}
Eq.~\ref{eq:sa_match} requires the layer-normalized hidden states 
of reference views, $\tilde{\mathbf{h}}_{\text{ref}}(q)$. We cache 
these by storing the output of the pre-attention LayerNorm 
($\text{LN}(\mathbf{h})$) during the reference pass, then applying 
L2 normalization along the feature dimension when computing 
cosine similarity. The same operation is applied to target view 
hidden states $\tilde{\mathbf{h}}_i(p)$ during their forward pass.

\paragraph{Nearest reference selection.}
For each target view $i$, we select its nearest references by 
Euclidean distance between camera centers:
\begin{equation}
  \mathcal{V}^{\text{near}}_i = \text{top-}k\!\left(
    \{ \|\mathbf{c}_i - \mathbf{c}_v\|_2 \}_{v \in \mathcal{V}_{\text{ref}}}
  \right),
  \label{eq:nearest_ref}
\end{equation}
where $k$ matches the per-batch reference count. Token correspondence 
in Eq.~\ref{eq:sa_match} is then computed only against tokens from 
$\mathcal{V}^{\text{near}}_i$, rather than the entire reference set. 
This local restriction reduces spurious matches between viewpoints 
that share little visual overlap.

\paragraph{Application scope.}
Self-attention alignment is applied at all \texttt{attn1} layers 
across all U-Net resolutions (down, mid, and up blocks).

\subsection{Cross-Attention Alignment: Detailed Operations}
\label{app:maa_ca_align}

\paragraph{Edit-relevant tokens.}
Given an edit prompt, we tokenize it with the CLIP tokenizer and 
extract all token indices except special tokens (BOS, EOS, PAD). 
This yields $T$, the set of edit-relevant token indices used in 
both the 3D attention field construction and the value projection 
$\mathbf{V}_T = W_V\, E_{\text{text}}[T]$.

\paragraph{Per-timestep 3D field reconstruction.}
The 3D attention field $M^{(t)}_{\text{3D}}$ is rebuilt at every 
denoising timestep $t \geq t_{\text{start}}$, since both the 
reference set and the attention maps $A^{(t)}_v$ change across 
timesteps. We use $t_{\text{start}} = 500$ in the standard 
1000-step schedule, focusing the alignment on the structure-forming 
phase of denoising. The full inverse-splatting formulation is 
given in Appendix~\ref{app:inverse_splat}.

\paragraph{Application scope.}
Cross-attention alignment is applied at the $32 \times 32$ 
attention resolution, where attention maps carry sufficient 
spatial detail to benefit from geometric alignment. At other 
resolutions ($16 \times 16$, $8 \times 8$, and the $64 \times 64$ 
layers without dense maps), the standard cross-attention is 
retained.

\paragraph{Aggregation across heads, tokens, and layers.}
Within the targeted resolution, the per-pixel cross-attention 
maps from reference views are averaged over attention heads, 
edit-relevant tokens, and U-Net layers before being lifted into 
the 3D attention field, yielding a single scalar field per 
Gaussian primitive. This aggressive reduction differs from 
VcEdit~\cite{wang2024vcedit}, which preserves head, token, and 
layer dimensions throughout. We find that the compact ACP view 
set provides sufficient angular diversity for the averaged field 
to remain well-conditioned, while the reduction yields substantial 
computational savings.

\subsection{Differences from Prior Work}
\label{app:maa_differences}

We summarize the design choices in MAA that differ from the 
methods we build upon:

\paragraph{No epipolar feature injection.}
While DGE~\cite{chen2024dge} additionally incorporates 
epipolar-constraint-based feature injection between distant 
COLMAP views, we omit this component for two reasons. First, 
ACP-selected reference views are spatially close enough to one 
another that the standard token correspondence 
(Eq.~\ref{eq:sa_match}) suffices. 
Second, the epipolar constraint computation introduces  
runtime overhead, which would offset the efficiency gains from 
our compact view set.

\paragraph{Reference-only cross-attention source.}
We collect $A^{(t)}_v$ only from reference views (approximately 
$20\%$ of the editing set), whereas VcEdit~\cite{wang2024vcedit} 
collects from all views via $K$ separate U-Net forward passes. 
This is enabled by ACP's diverse viewpoint coverage: reference 
views alone provide sufficient angular diversity for the 3D 
attention field to be well-conditioned. The reduction yields 
roughly $K \times$ savings in the cross-attention map collection 
step.

\paragraph{Single-resolution cross-attention replacement.}
We apply the alignment only at $32 \times 32$, while VcEdit 
processes all attention layers. We find the single-resolution 
variant sufficient for ACP-selected views and substantially less 
expensive.

\subsection{Joint Forward Pass for Target Views}
\label{app:maa_joint_forward}

For each target view $i$ at timestep $t$, the U-Net forward pass 
proceeds through a stack of transformer blocks, each containing 
a self-attention (\texttt{attn1}), cross-attention (\texttt{attn2}), 
and feed-forward sublayer. Within a single block:
\begin{enumerate}
  \item LayerNorm produces $\tilde{\mathbf{h}}_i$;
  \item \texttt{attn1} output is replaced via 
  Eq.~\ref{eq:sa_replace}, followed by residual addition;
  \item LayerNorm again, then \texttt{attn2} output is replaced 
  via Eq.~\ref{eq:ca_replace} (at the targeted resolution), 
  followed by residual addition;
  \item Feed-forward and final residual.
\end{enumerate}
Both replacements occur within the same forward pass, so the two 
alignments are applied jointly rather than in separate passes.

\section{Inverse Splatting for 3D-Consistent Maps}
\label{app:inverse_splat}

Our pipeline uses inverse splatting---lifting per-view 2D maps 
onto Gaussian primitives and re-rendering them into arbitrary 
views---as a unified mechanism for two purposes: (i) constructing 
a 3D ROI mask for localized editing, and (ii) sharing 
cross-attention maps across views for spatial consistency. Both 
follow the same \emph{lift $\rightarrow$ aggregate $\rightarrow$ 
re-render} pattern, differing only in what 2D signal is lifted.

\subsection{General Formulation}
Let $F_v \in \mathbb{R}^{HW \times C}$ denote a per-pixel 2D 
feature map from view $v$ (either a binary segmentation mask with 
$C{=}1$ or a scalar cross-attention field with $C{=}1$ after 
aggregation). For each Gaussian primitive $g$, we project it into 
view $v$ to obtain its 2D footprint $\pi_v(g)$ and read off the 
corresponding feature value. We then aggregate values across all 
contributing views via a count-weighted average:
\begin{equation}
  F_{\text{3D}}(g, c) = \frac{
    \sum_{v \in \mathcal{V}}\, \mathbb{1}[g \in \text{view}_v]\, 
    F_v(\pi_v(g), c)
  }{
    \sum_{v \in \mathcal{V}}\, \mathbb{1}[g \in \text{view}_v] + \epsilon
  },
  \label{eq:inverse_splat_general}
\end{equation}
where $\mathbb{1}[g \in \text{view}_v]$ indicates whether $g$ is 
visible in view $v$. We deliberately use a simple count-average 
rather than transmittance-weighted accumulation to avoid biasing 
the 3D field toward views where $g$ happens to be foremost; this 
yields more stable aggregation across diverse viewpoints. The 
resulting per-Gaussian field $F_{\text{3D}}$ is re-rendered into 
any target view $i$ via the standard differentiable Gaussian 
rasterizer, which performs the usual alpha-compositing along each 
ray:
\begin{equation}
  \hat{F}_i(p, c) = \text{GSRender}\!\bigl(F_{\text{3D}}(\cdot, c),\;
  \text{cam}_i,\; p\bigr).
  \label{eq:rerender_general}
\end{equation}
That is, the lifting step (Eq.~\ref{eq:inverse_splat_general}) 
uses count-averaging, while the re-rendering step uses standard 
3DGS alpha-compositing.

\subsection{Application 1: 3D ROI Masking}
To restrict editing to the target object, we run 
LangSAM~\cite{langsam} on $n$ rendered COLMAP views (default 
$n = 8$, sampled uniformly along the trajectory) to obtain binary 
segmentation masks $F_v = M_v \in \{0, 1\}^{HW}$. Applying 
Eq.~\ref{eq:inverse_splat_general} with $C{=}1$ yields a 
per-Gaussian ROI score; thresholding at $0.3$ produces the 3D 
ROI mask $\mathcal{M}_{\text{ROI}} \subseteq \mathcal{G}$. This 
mask serves two roles: ACP uses it to compute attention statistics 
for view selection (Sec.~\ref{sec:acp}), and during 3DGS 
fine-tuning it constrains the optimization to only the Gaussians 
belonging to the ROI.

\subsection{Application 2: Cross-Attention Alignment}

\paragraph{3D attention field construction.}
During multi-view editing, we set $F_v = \bar{A}^{(t)}_v$, where 
$\bar{A}^{(t)}_v \in \mathbb{R}^{HW}$ is the cross-attention 
probability of reference view $v$ at denoising timestep $t$, 
averaged over attention heads, edit-relevant tokens $T$, and 
U-Net layers at the targeted resolution. Applying 
Eq.~\ref{eq:inverse_splat_general} yields a shared 3D attention 
field $M^{(t)}_{\text{3D}}(g)$ on the Gaussian primitives. 
Re-rendering via Eq.~\ref{eq:rerender_general} produces 
geometrically consistent 2D attention maps $\hat{A}^{(t)}_i$ for 
every target view, which are then expanded back to the 
$|T|$-token dimension before the cross-attention replacement 
(Eq.~\ref{eq:ca_replace}).

\paragraph{Cross-attention replacement.}
During the target-view U-Net forward pass, the standard 
cross-attention output $\text{softmax}(QK^\top/\sqrt{d})\, V$ is 
replaced by:
\begin{equation}
  \mathbf{o}^{\text{cross}}_i = \hat{A}^{(t)}_i\, \mathbf{V}_T,
  \label{eq:cross_replace}
\end{equation}
where $\mathbf{V}_T = W_V\, E_{\text{text}}[T]$ contains the 
value projections of the edit-relevant text tokens.

\begin{figure*}[t]
    \centering
    \includegraphics[
        width=0.80\textwidth,
        clip,
        trim=0cm 1cm 0cm 1cm
    ]{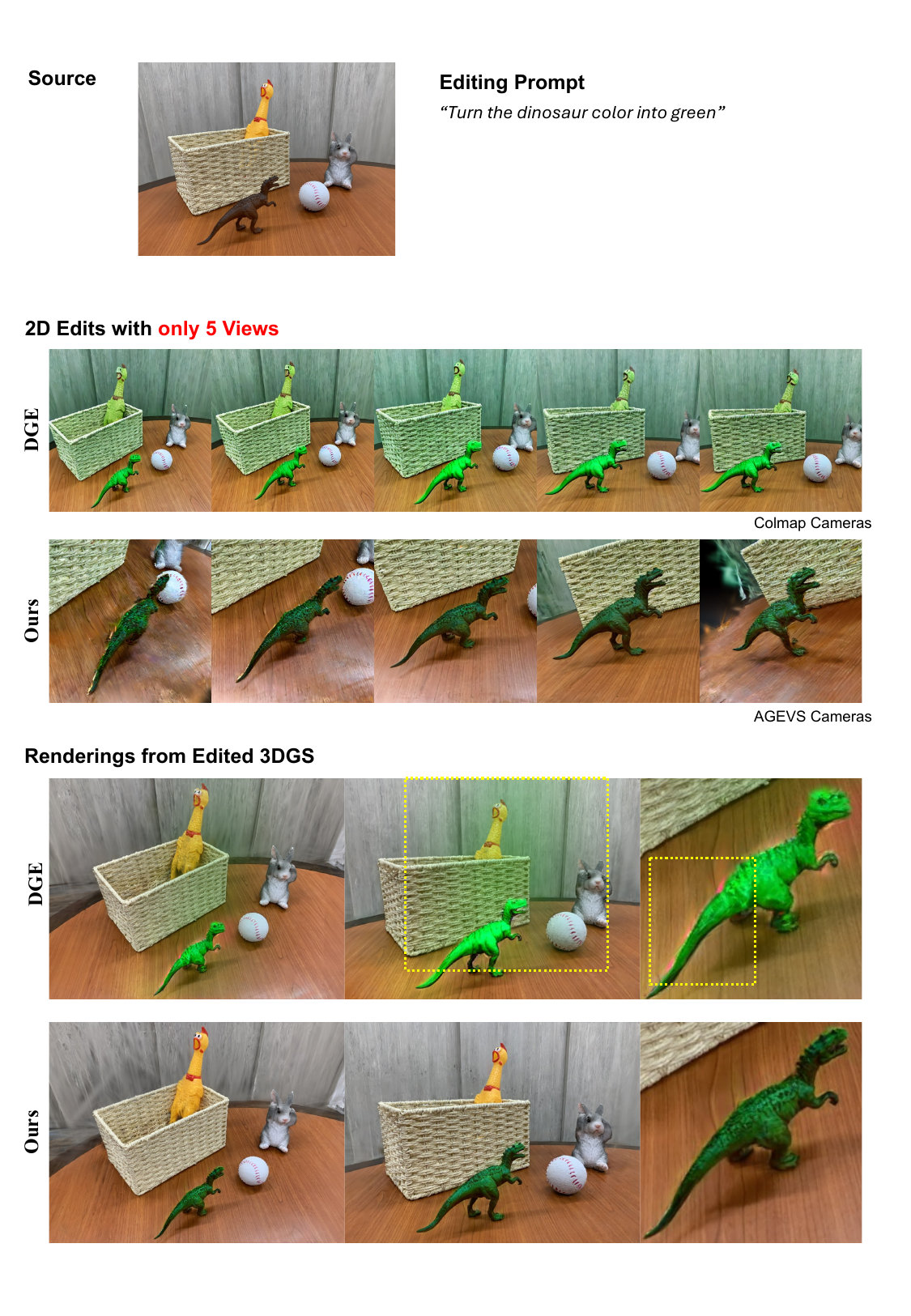}
    \caption{
    \textbf{Editing with only 5 views.}
    DGE selects 5 random COLMAP cameras, leading to global color bleeding and
    rendering artifacts (red spiky Gaussians, blurry green clouds). Our ACP
    cameras focus on the target object, producing clean edits even with very few views.
    }
    \label{fig:app-few_edits}
\end{figure*}

\section{More Experiments}

\subsection{Consistent Multi-view Edits}
Fig.~\ref{fig:app-multiview_edits} compares the multi-view 2D editing results of DGE (using COLMAP cameras) and our method (using ACP cameras) on two challenging editing prompts. 
Maintaining multi-view consistency under instance-level transformations that introduce previously nonexistent geometry is among the hardest editing tasks—for example, transforming a man into a clown requires generating a round red nose whose shape differs substantially from the original.

In the \textit{pooh $\rightarrow$ robot} example, DGE produces a nose-like protrusion that appears only from certain viewpoints, whereas our method renders the feature consistently across all viewing directions.
In the \textit{man $\rightarrow$ clown} example, DGE generates two to three redundant noses per viewpoint. This artifact stems from epipolar-constraint-based feature injection: when depth estimates are inaccurate between distant COLMAP views, conflicting information about nose placement is propagated to neighboring views, resulting in duplicated structures.
In contrast, our ACP cameras are spaced closely enough to maintain reliable depth correspondence while still providing sufficient viewpoint diversity. Combined with similarity-based self-attention sharing and 3D-guided cross-attention sharing (Sec.~\ref{sec:maa}), the spatial information about where the nose should appear is accurately propagated, producing a single coherent sphere across all views.

\subsection{Editing with a Few Views}
Existing methods edit a subset of COLMAP cameras, so reducing the number of editing views significantly degrades multi-view consistency. Fig.~\ref{fig:app-few_edits} compares DGE and our method on the \textit{room} scene with the prompt ``Turn the dinosaur color into green,'' each using only 5 editing views.

DGE randomly selects 5 views from the COLMAP camera set, which tend to be distant and cover similar viewpoints. As a result, the 2D edits turn not only the dinosaur but the entire image green, and the limited angular diversity means only a narrow range of viewpoints is supervised. When the edited images are used to fine-tune 3DGS, the renderings exhibit red, spiky Gaussian artifacts around the dinosaur's tail and blurry green clouds throughout the scene.

Our method generates 5 ACP cameras centered on the dinosaur, covering nearly all viewpoints from which the target object is visible. Despite the small number of views, this provides sufficient supervision for 3DGS fine-tuning, and the resulting renderings are clean and free of spurious Gaussian artifacts.

\subsection{Comparison on Editing Time}
To compare the computational cost of 3DGS editing, we measure the runtime of each method on the \textit{room} scene with the prompt ``Turn the dinosaur color into green.'' GSEditor and VcEdit use all 31 COLMAP cameras following their original implementations, while DGE and our method are evaluated with 20 and 5 editing views.

As shown in Table~\ref{tab:runtime_breakdown}, GSEditor and VcEdit require approximately 17--20 minutes in total, dominated by multi-view 2D image editing, and yield relatively low CLIPsim scores. DGE and our method are substantially faster, with total times roughly proportional to the number of editing views. 

Our method shows a slight decrease in CLIPsim when reducing from 20 to 5 views, reflecting the expected trade-off between coverage and speed. Notably, DGE at 5 views shows an unusually high CLIPsim score (15.30). This is an outlier: as shown in Fig.~\ref{fig:app-few_edits}, DGE's editing turns the entire scene green, and since CLIPsim evaluates the whole image, the metric mistakenly considers the global color shift as a successful edit from ``brown dinosaur'' to ``green dinosaur.'' This score should therefore be treated with caution.

% -----------------------------------------------------------------------------
% Appendix: More Qualitative Comparisons, Prompt Details, and User Study Interface
% -----------------------------------------------------------------------------

\section{Evaluation Prompt Details}
\label{app:prompt_details}

Table~\ref{tab:prompt_details} lists the editing prompts, segmentation prompts,
and CLIPsim source/target prompts used for evaluation.

% -----------------------------------------------------------------------------
% Appendix: Prompt Details and User Study Interface
% -----------------------------------------------------------------------------

\section{Evaluation Prompt Details}
\label{app:prompt_details}

Table~\ref{tab:prompt_details} lists the editing prompts, segmentation prompts,
and CLIPsim source/target prompts used for evaluation.

\begin{table*}[t]
\centering
\caption{
Editing prompts, segmentation prompts, and CLIPsim prompts used for evaluation.
}
\label{tab:prompt_details}

\scriptsize
\setlength{\tabcolsep}{3pt}
\renewcommand{\arraystretch}{1.08}

\begin{tabular}{
    >{\raggedright\arraybackslash}p{1.35cm}
    >{\raggedright\arraybackslash}p{3.25cm}
    >{\raggedright\arraybackslash}p{2.45cm}
    >{\raggedright\arraybackslash}p{4.25cm}
    >{\raggedright\arraybackslash}p{4.25cm}
}
\toprule
\textbf{Scene} &
\textbf{Editing Prompt} &
\textbf{Segmentation Prompt} &
\textbf{CLIPsim Source Prompt} &
\textbf{CLIPsim Target Prompt} \\
\midrule

covered\_desk & Make pooh look like a panda & pooh & pooh & panda \\
covered\_desk & Make pooh look like a robot & pooh & pooh & robot \\
covered\_desk & Change pooh's sweater color to blue & pooh's red sweater & pooh's red sweater & pooh's blue sweater \\

room & Change the baseball to a solid gold ball & white baseball & white baseball & solid gold ball \\
room & Change the rubber chicken's color to red & yellow rubber chicken & yellow rubber chicken & red rubber chicken \\
room & Change the dinosaur figure color to green & brown dinosaur figure & brown dinosaur figure & green dinosaur figure \\
room & Change the dinosaur figure material to shiny metal & brown dinosaur figure & brown dinosaur figure & shiny metallic dinosaur figure \\
room & Give the rabbit figure a pair of small sunglasses & grey rabbit figure & grey rabbit figure & grey rabbit figure wearing small sunglasses \\

blue\_sofa & Change the sunglasses to red frames & black sunglasses frames & black sunglasses frames & bright red sunglasses frames \\
blue\_sofa & Change the plush toy's color to pink & yellow plush toy & yellow plush toy & pink plush toy \\
blue\_sofa & Change the speaker to a shiny gold texture & silver JBL speaker & silver JBL speaker & shiny gold JBL speaker \\
blue\_sofa & Change the remote control color to yellow & remote control & white remote control & yellow remote control \\

face & Turn the man into a clown & man & man & clown \\
face & Turn the man into a grandma & man & man & grandma \\
face & Change the jacket material to blue denim & entire fleece jacket & man wearing a jacket with textured grey speckled fleece fabric & man wearing a jacket with classic blue denim twill fabric \\
face & Turn the man into Spider-Man with a mask & man & man & Spider-Man with a mask \\
face & Change his eye color to blue & face & man with brown eye irises & man with blue eye irises \\

bear & Change the material of the bear statue to clear ice & entire bear statue & bear statue with grey stone texture & bear statue with transparent ice texture with reflections \\
bear & Change the material of the bear statue to bronze & entire bear statue & bear statue with grey stone texture & bear statue with aged bronze metal texture \\
bear & Change the bear statue into a polar bear & entire bear statue & bear statue with grey stone texture & polar bear \\

\bottomrule
\end{tabular}
\end{table*}

% \FloatBarrier

\section{User Study Interface}
\label{app:user_study_interface}

\begin{figure*}[t]
    \centering
    \includegraphics[
        width=0.98\textwidth,
        clip,
        trim=0cm 2cm 7cm 0cm
    ]{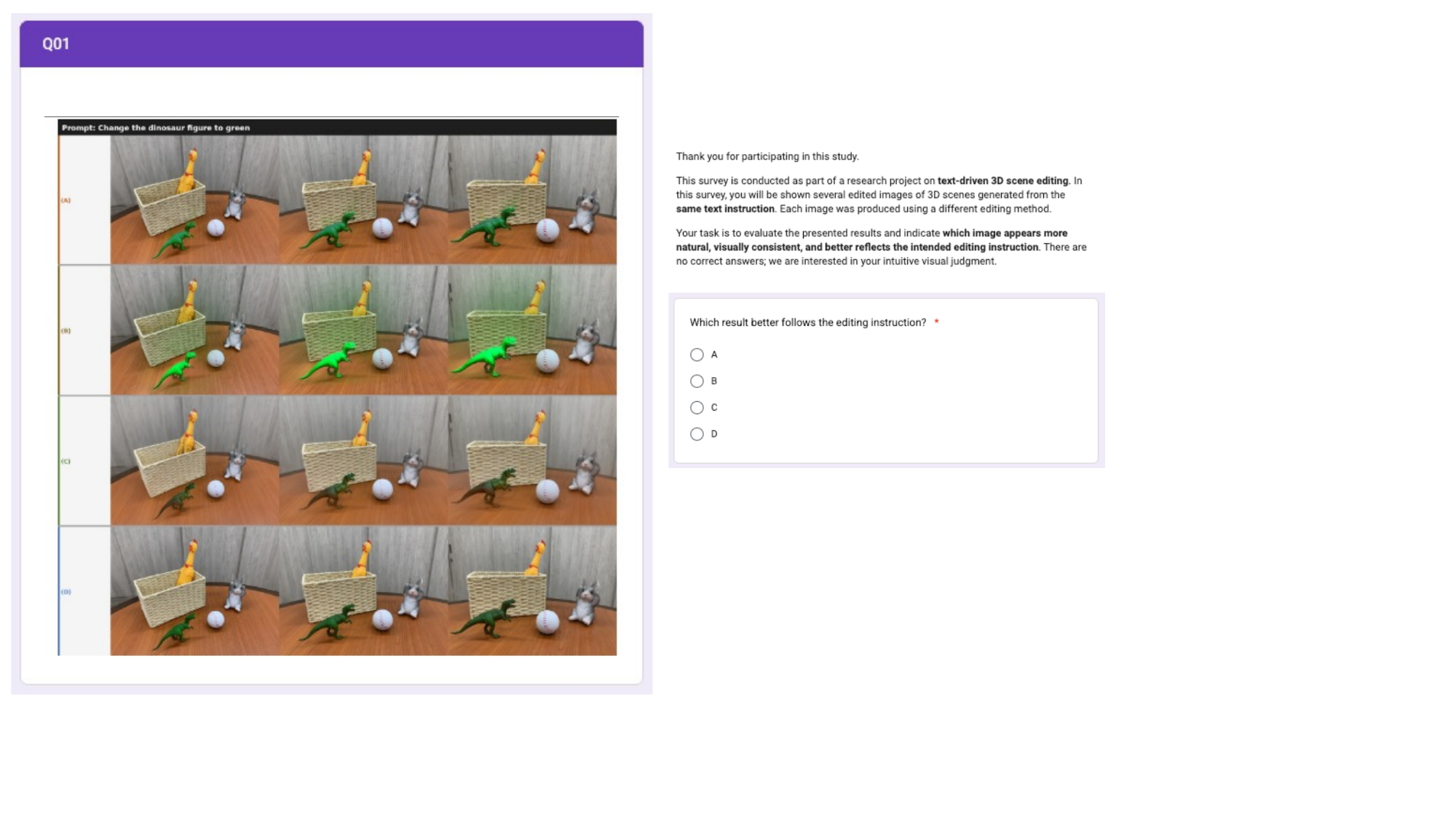}
    \caption{
    \textbf{User study interface.}
    Participants are shown four anonymized editing results side by side and asked
    to select the best one based on instruction fidelity, multi-view consistency,
    and editing locality.
    }
    \label{fig:app-user_study}
\end{figure*}

Figure~\ref{fig:app-user_study} shows the interface used in our user study.
Participants were shown four anonymized editing results side by side and asked to
select the best result according to instruction fidelity, multi-view consistency, and
editing locality.

\FloatBarrier

\end{document}